\documentclass{article}

\usepackage[preprint]{neurips_2022}

\usepackage[utf8]{inputenc} 
\usepackage[pagebackref=true]{hyperref}       
\usepackage{url}            
\usepackage{booktabs}       
\usepackage{amsfonts}       
\usepackage{nicefrac}       
\usepackage{microtype}      
\usepackage{xcolor,wrapfig}         
\usepackage{algorithm}
\usepackage{algpseudocode}
\usepackage[nohints]{minitoc}

\usepackage{natbib,enumitem}
\newif\ifneurips
\neuripstrue

\usepackage{float}

\usepackage{amsmath, mathtools, amsthm, amssymb, algorithm, algpseudocode, times, yhmath}

\theoremstyle{plain}
\newtheorem{theorem}{Theorem}
\newtheorem{lemma}[theorem]{Lemma}

\theoremstyle{definition}
\newtheorem{definition}[theorem]{Definition}
\newtheorem{assumption}{Assumption}

\theoremstyle{remark}
\newtheorem{remark}{Remark}

\renewcommand{\Pr}{\field{P}}
\newcommand{\E}{\field{E}}
\newcommand{\Var}{\mathrm{Var}}

\newcommand{\be}{\boldsymbol{e}}

\newcommand{\bx}{\boldsymbol{x}}

\newcommand{\bX}{\boldsymbol{X}}
\newcommand{\bY}{\boldsymbol{Y}}

\newcommand{\bw}{\boldsymbol{w}}

\newcommand{\bK}{\mathbb{K}}

\newcommand{\bv}{\boldsymbol{v}}

\newcommand{\sD}{\mathcal{D}}

\newcommand{\sF}{\mathcal{F}}

\newcommand{\sL}{\mathcal{L}}
\newcommand{\sM}{\mathcal{M}}
\newcommand{\sN}{\mathcal{N}}

\newcommand{\sR}{\mathcal{R}}

\newcommand{\sX}{\mathcal{X}}
\newcommand{\sY}{\mathcal{Y}}

\newcommand{\bfB}{\mathbf{B}}

\newcommand{\argmin}{\mathop{\mathrm{argmin}}}

\newcommand{\range}{\mathop{\mathrm{range}}}

\newcommand{\field}[1]{\mathbb{#1}}

\newcommand{\R}{\field{R}}

\newcommand\inn[2]{ \left\langle {#1} , {#2} \right\rangle }

\newcommand{\norm}[1]{\left\|{#1}\right\|}
\newcommand{\ltwonorm}[1]{\left\|{#1}\right\|_2}

\DeclarePairedDelimiter{\ceil}{\lceil}{\rceil}

\newcommand{\indicator}{\mathbf{1}}

\usepackage[most]{tcolorbox}
\newtcolorbox{idea}[1][]
{
colbacktitle=cyan,
colback=cyan!10,
arc=1pt,
boxrule=1pt,
title=#1 
}

\newtcolorbox{update}[1][]
{
colbacktitle=gray,
colback=gray!10,
arc=1pt,
boxrule=1pt,
title=#1 
}

\newtcolorbox{question}[1][]
{
coltitle=black,
colbacktitle=yellow,
colback=yellow!10,
arc=1pt,
boxrule=1pt,
title=#1 
}

\newtcolorbox{note}[1][]
{
coltitle=black,
colbacktitle=green,
colback=green!10,
arc=1pt,
boxrule=1pt,
title=#1 
}

\DeclarePairedDelimiterX{\innerproduct}[2]{\langle}{\rangle}{#1, #2}

\newcommand{\func}[3]{{#1} : {#2} \rightarrow {#3}}

\newcommand{\bfX}{\mathbf{X}}

\newcommand{\identity}{\mathbb{I}}

\newcommand{\ybound}{\boundony}
\newcommand{\sampleD}{t}
\newcommand{\K}{\kappa}
\newcommand{\linweight}{\lassobound}
\newcommand{\fspace}{\sF}
\newcommand{\linfunc}{\sF_{\mathrm{lin}}}
\newcommand{\blinfunc}{\sF_{\linweight}}
\newcommand{\sampD}{\sD_{\surveyset}}
\newcommand{\D}{\sD^*}
\newcommand{\valD}{\D}
\newcommand{\tol}{\epsilon}
\newcommand{\confidence}{\delta}
\newcommand{\xdomain}{\sX}
\newcommand{\yrange}{\sY}
\newcommand{\distF}{dist_{\D}}
\newcommand{\distltwo}{dist_{\sD}}

\newcommand{\lip}{\ell}
\newcommand{\Loss}{\sL}
\newcommand{\optimalf}{f^*}
\newcommand{\surveyf}{f_\surveyset}
\newcommand{\optimalsurveyf}{f^*_\surveyset}
\newcommand{\sizeS}{m}
\newcommand{\estimatevarS}{\hat{\gamma}_S}
\newcommand{\estimatevarD}{\hat{\gamma}_{\D}}
\newcommand{\standevnoise}{\sigma_\regnoise}
\newcommand{\varnoise}{\sigma_\regnoise^2}
\newcommand{\sdnoise}{\sigma_\regnoise}
\newcommand{\SurVerif}{\ensuremath{\texttt{SurVerify}}}
\newcommand{\PriVerif}{\ensuremath{(\alpha,\beta)-\texttt{PriVerify}}}
\newcommand{\PriVerifL}{\ensuremath{(\alpha)-\texttt{PriVerify}}}

\newcommand{\boundonx}{\zeta}
\newcommand{\boundony}{\tau}
\newcommand{\Sensitivity}{\Delta}
\newcommand{\dpepsilon}{\alpha}
\newcommand{\dpdelta}{\beta}
\newcommand{\dppair}{$(\dpepsilon,\dpdelta)$}
\newcommand{\coeff}{\boldsymbol{\theta}}
\newcommand{\surveyset}{S}
\newcommand{\response}{y}
\newcommand{\regnoise}{\eta}    
\newcommand{\covariate}{\mathbf{x}} 
\newcommand{\covariatei}{x}         
\newcommand{\covariatematrix}{\mathbf{X}}
\newcommand{\AlgoPriv}{\texttt{Priv-n-Pub}}
\newcommand{\AlgoReg}{\texttt{LASSO-SEN}}
\newcommand{\noisedist}{\sD_\noisevariable}
\newcommand{\noisevariance}{\mathbf{\Sigma}_\noisevariable}
\newcommand{\noisevariable}{\mathbf{q}}     
\newcommand{\noboldnoisevariable}{q}        
\newcommand{\noisematrix}{\mathbf{Q}}
\newcommand{\xvariance}{\mathbf{\Sigma}_\covariate}
\newcommand{\coeffhat}{\hat{\coeff}}
\newcommand{\basecovardummy}{\mathbf{\hat{\Gamma}}}
\newcommand{\predcovardummy}{\boldsymbol{\hat{\gamma}}}
\newcommand{\noisycovariate}{\mathbf{z}}    
\newcommand{\zvariance}{\mathbf{\Sigma}_\noisycovariate}
\newcommand{\noisycovariatematrix}{\mathbf{Z}}
\newcommand{\lassobound}{R}
\newcommand{\constant}{c}
\newcommand{\lowerrealpha}{\alpha_\ell}

\newcommand{\responsevector}{\mathbf{Y}}
\newcommand{\wvariance}{\mathbf{\Sigma_w}}
\newcommand{\regnoisevector}{\boldsymbol{\regnoise}}

\newcommand{\coeffdist}{\psi_{DP}}
\newcommand{\dpsurveyset}{\surveyset_{DP}}

\newcommand{\acsincome}{ACS\textunderscore Income }
\newcommand{\laplace}{\mathrm{Lap}}
\newcommand{\uniform}{\mathrm{Unif}}
\newcommand{\normal}{\sN}

\title{Testing Credibility of Public and Private Surveys through the Lens of Regression}

\author{%
  Debabrota Basu\\
  \'Equipe Scool, Univ. Lille, Inria,\\ 
  CNRS, Centrale Lille, UMR 9189- CRIStAL\\ 
  F-59000 Lille, France\\
  \And
  Sourav Chakraborty \\
  Indian Statistical Institute \\
  Kolkata, India\\
  \And
  Debarshi Chanda \\
  Indian Statistical Institute \\
  Kolkata, India\\
  \And
  Buddha Dev Das \\
  Indian Statistical Institute \\
  Kolkata, India\\
  \And
  Arijit Ghosh \\
  Indian Statistical Institute \\
  Kolkata, India\\
  \And
  Arnab Ray \\
  Indian Statistical Institute \\
  Kolkata, India\\
}

\begin{document}

\maketitle
\doparttoc 
\faketableofcontents 
\begin{abstract}
Testing whether a sample survey is a credible representation of the population is an important question to ensure the validity of any downstream research. While this problem, in general, does not have an efficient solution, one might take a task-based approach and aim to understand whether a certain data analysis tool, like linear regression, would yield similar answers both on the population and the sample survey. 
In this paper, we design an algorithm to test the credibility of a sample survey in terms of linear regression. In other words, we design an algorithm that can certify if a sample survey is good enough to guarantee the correctness of data analysis done using linear regression tools. 
Nowadays, one is naturally concerned about data privacy in surveys. Thus, we further test the credibility of surveys published in a differentially private manner. 
Specifically, we focus on Local Differential Privacy (LDP), which is a standard technique to ensure privacy in surveys where the survey participants might not trust the aggregator. We extend our algorithm to work even when the data analysis has been done using surveys with LDP. In the process, we also propose an algorithm that learns with high probability the guarantees a linear regression model on a survey published with LDP.
Our algorithm also serves as a mechanism to learn linear regression models from data corrupted with noise coming from any subexponential distribution. We prove that it achieves the optimal estimation error bound for $\ell_1$ linear regression, which might be of broader interest.
We prove the theoretical correctness of our algorithms while trying to reduce the sample complexity for both public and private surveys. We also numerically demonstrate the performance of our algorithms on real and synthetic datasets.

\end{abstract}

%

\section{Introduction}
Socio-economic surveys are conducted internationally to gather data on population characteristics for various purposes, including demographic and economic analyses, educational planning, poverty studies, exit poll analysis, and assessing progress towards national objectives~\citep{Groves2011survey,Kenny2021_US-Census}. These recurring surveys are crucial for monitoring and evaluating the impact of different policies over time~\citep{banerjee2020governance}.
However, in practice, they are often conducted using logistically constrained data collection methods and are used as benchmarks over decades to validate research hypotheses~\citep{salant1994conduct,statscanada}. It is essential to determine whether the collected survey sample accurately represents the population to ensure the validity of subsequent research. This concern is a well-known problem in both statistics and computer science, often described in the latter field as the challenge of determining the closeness of two distributions~\citep{Batu2000_equivalence,CanonneTopicsDT2022}. Unfortunately, solutions to this problem are often inefficient, typically requiring an exponential number of new samples.

The primary purpose of such surveys is to enable deduction-based analyses to identify patterns that inform future research and policy-making~\citep{heeringa2017applied,statscanada}.
Although determining whether a sample survey accurately represents the entire population may currently be beyond our reach, it is crucial to test the accuracy of the deductions made from such surveys. Specifically, if the deduction techniques belong to a well-established class of tools,\textit{ we should be able to certify that any conclusions drawn using these tools from the given survey are valid}.

One widely used and interpretable method for conducting these analyses on survey data is fitting a linear regression model. For instance, \cite{Balia2008_survey} employs data from the British Health and Lifestyle Survey (1984–1985) and its longitudinal follow-up in May 2003 to demonstrate a strong association between mortality and socio-economic status. 
In this paper, we ask the following question:



\begin{center}
    \textit{Can we certify that deductions made using linear regression models on a given sample survey data would yield similar results if applied to the entire population}? 
\end{center}

Conducting a sample survey with a large number of data points is complex and expensive, often leading to compromised data quality. However, the expectation is that by collecting a smaller number of additional high-quality data points, the overall quality can be validated. Therefore, the main approach to address the above question involves using a small amount of high-quality additional data, alongside the original sample survey data, to certify the credibility of survey data in relation to linear regression models. \textit{The main goal is to design an efficient algorithm that optimizes both the running time and the sample complexity (the new samples necessary)}.

In the modern age, concerns about data privacy in surveys are paramount. Due to the risk of privacy breaches in data-driven applications that involve personal or confidential data, survey aggregators have taken significant steps to protect respondent's privacy~\citep{Plutzer_Survey,Connors_Survey}, typically by removing sensitive information such as names, addresses, and contact numbers. However, it has been shown that these de-identification methods do not fully protect against intentional adversarial re-identification attacks~\citep{DBLP:conf/pods/DinurN03, Henriksen2016rei,Wood2018differential}. A promising solution to this challenge is the use of Differential Privacy (DP), which provides plausible deniability for individuals by introducing random noise into the data. Two main models have emerged for implementing differential privacy: the central model~\citep{Dwork2006_DP-survey} and the local model~\citep{MengmengYang2020}. In the centralized DP, additive noise is added while processing an entire central database. 
However, this approach requires users to trust the database curator to uphold their privacy. In contrast, the local model operates on a client-server basis, where each individual manages their own data and shares it with a server using differentially private mechanisms. The server then aggregates these randomized responses, ensuring plausible deniability for each user while enabling accurate data interpretation. These DP mechanisms are increasingly applied to survey sampling with the new privacy regulations across the globe~\citep{Kenny2021_US-Census,evans2022differentially}. 

Thus, in the context of the main question of this paper, we extend our study to consider scenarios where the sample data is collected, released, and used after being ``hidden" using local DP. This adds an additional layer of complexity to the problem, as designing an efficient algorithm becomes even more challenging without access to the original data sample.

Existing literature includes several studies on privacy-preserving linear regression~\citep{,dandekar2018differential, Diff_private_Simple_regression_AlabiMSSV22,ICLR23_Diff_private_LinReg}, and more broadly, private convex optimization using central differential privacy (central-DP) as the privacy framework~\citep{iyengar2019towards}. However, due to the severely limited functionality of data adhering to local DP, even widely used methods like linear regression analysis have not been extensively explored in this context. This raises another important question:
\begin{center}
    \textit{Does there exist an efficient way to learn a linear regression model on survey data satisfying local privacy guarantees?}
\end{center}

\subsection{Problem Formulation}\label{sec:formulation}
Before we answer both the aforementioned questions, we first formally propose them here.

\paragraph{Testing Credibility of Surveys.} 
Typically, after a sampling-based study is designed, sample surveys are collected from an underlying population. 
Following the structure of the majority of socio-economic surveys, we assume that the survey data $\surveyset$ consists of tabular numeric covariates and a scalar response variable. This means that $\surveyset$ contains data points of the form $(\covariate,\response)$, where the covariates $\covariate \in \R^d$ and the response variable $\response \in \R$.

We denote by $\D$ the distribution of the $(\covariate,\response)$ tuples of the whole population. 
If the dataset $\surveyset$ was obtained after perfect sampling techniques, i.e. by drawing independent samples from an unknown distribution $\D$, then one would call the survey data $\surveyset$ to be a credible representation of the population. 
But due to various limitations, the dataset $\surveyset$ collected might be obtained by drawing samples from some other distribution $\sampD$. So the question about how credible is $\surveyset$ as a representation of the population boils down to understanding the distance between the two unknown distributions $\D$ and $\sampD$.  We will call $\D$ to be the true distribution and $\sampD$ to be the sample distribution.
Estimating the distance between two high-dimensional distributions is very inefficient, and hence, impractical~\citep{Canonne:Survey:ToC,CanonneTopicsDT2022}. 

Samples collected from a survey are typically used for various data interpretation and deduction tasks, e.g. regression, classification etc.
In all these cases, one aims to find a model from a given model class, say $\fspace$, that minimises a task-specific loss function. For example, for linear regression, we aim to find out the coefficients that minimise the square loss over the survey data.  If $\Loss:\R^2 \to \R$ is the loss function, then the model learnt from the survey set $\surveyset$ is

$$f_{\surveyset} \triangleq \argmin_{f \in \fspace} \frac{1}{m} \sum_{(\covariate,\response) \in \surveyset}\Loss (f(\covariate),\response).$$

To validate the credibility of a survey data, we propose to test whether the model $f_{\surveyset}$ derived from the survey data $\surveyset$ matches the model $\optimalf$ that would have been derived had the dataset $\surveyset$ been a credible representation of the population. 
 
$$\optimalf \triangleq \argmin_{f \in \fspace} \E_{(\covariate,\response)\sim \D} \Loss(f(\covariate), \response).$$

We will assume that we have access to a small sample set, called the validation dataset, obtained by drawing independent samples from the true distribution $\D$. 

In this paper, our goal is to validate the quality of the survey data $\surveyset$ by estimating the distance of $f_{\surveyset}$ from $\optimalf$.
Depending on the problems, different metrics have been proposed to quantify the closeness of functions evaluated over distributions~\citep{Gibbs2002_metrics}. In this paper, we use the distributional $\ell_2$ distance to quantify the closeness of linear regression models. 
\begin{definition}\label{Definition: Statistical distance between functions}
    \textbf{(Distributional $\ell_2$-Distance between Functions).}  Let $f$ and $g$ be real-valued functions on $\R^{d}$, and $\sD$ be a distribution on $\R^{d}$. The distributional $\ell_2$-distance between $f$ and $g$ on $\sD$ is 
        $\distltwo(f,g) \triangleq \sqrt{\E_{\covariate \sim \sD} (f(\covariate)-g(\covariate))^2}.$
\end{definition}

Given this distance measure we introduce the notion of distance between the survey data $\surveyset$ and $\valD$.


\begin{definition}
    A survey data $\surveyset$ is said to be $\K$-close to a credible data set $\valD$ with respect to $\fspace$ if $\distF(f_{\surveyset}, \optimalf) \leq \K. $
    Similarly, if $\distF(f_{\surveyset}, \optimalf) \geq \K$ the survey data $\surveyset$ is said to be $\K$-far from being a credible data set with respect to $\fspace$.
\end{definition}

The problem of testing the credibility of the survey data $\surveyset$ is checking whether $\surveyset$ is $\K$-close to a credible data set with respect to $\fspace$ using as small number of samples as possible for the validation set.



In this paper, we focus on the bounded linear models learned through a linear regression procedure, which is still one of the widely used models for socio-economic data~\citep{stanley2008meta}. Please refer to Section~\ref{Section : Linear Regression} for a more detailed discussion on linear regression models. 
In this paper, we only consider the linear regression models with bounded coefficients. 

\begin{definition}
    A set of bounded linear functions $\blinfunc$ is defined as $\blinfunc = \{f(\covariate) = \inn{\coeff}{\covariate}|\norm{\coeff}_1 \leq \linweight\}$.
\end{definition}

\paragraph{Learning Linear Regression on Private Surveys.} This brings us to a natural question about the learnability of a bounded linear regression model from sampled data. In literature, this problem has been widely studied in multiple setups, e.g. independent data, dependent data, noisy data, etc.~\citep{LDP_Regression_AOS,ouhamma2021stochastic,della2023online}. We can leverage this analysis while working on publicly available survey data and validation data.

But, due to the privacy concerns in surveys, new surveys are published in a differentially private manner~\citep{Kenny2021_US-Census}. 
Differential Privacy (DP) is a statistical technique that hinders the identification of an individual data point while looking into a published survey or any aggregated statistics on it~\citep{Dwork2014_DP-book}. Here, we focus on locally DP publication of surveys where one reports a privatized version of data $(\tilde{\covariate},\response)$ than the true one $(\covariate,\response)$.
In order to proceed with our credibility testing scheme, we now have to learn a linear regression model on this locally private survey data, say $\dpsurveyset$.
We investigate the question:
\textit{How many samples do we need to learn an $\epsilon$-correct linear regression model from an $(\alpha, \beta)$-local DP survey data?}


\subsection{Our Contributions}
We make the following assumptions for the rest of this paper.
\begin{assumption}[\textbf{Linear Regression Model}]\label{Assumption : Assumptions on the linear regression model}
We assume that the data is generated from a linear regression model $\response = \inn{\coeff}{\covariate}+\regnoise$ such that
\begin{itemize}[leftmargin=*]
    \item \emph{Homoscedasticity of errors: } The noise $\regnoise$ has constant variance, i.e. $\Var[\regnoise\mid\covariate] = \varnoise$.
    \item \emph{Non-correlation of errors: } The noise $\regnoise$ is uncorrelated with $\covariate \in \R^d$ and independent across observations.
\end{itemize}
\end{assumption}

\begin{assumption}[\textbf{Bounded Variables}]\label{Assumption : Bounded values for response variable} 
     We assume that the response variable $\response \in [-\boundony,\boundony]$ for $\boundony > 0$, and the covariates satisfy $|\covariate_i| \leq \boundonx$ for all $i \in [d+1]$ and $\boundonx >0$.
\end{assumption}
Under these assumptions, we elaborate the main contributions of this paper.

\noindent\textbf{1. Task-specific Credibility Testing.}  First, we introduce the formulation of task-specific credibility testing of datasets (Section~\ref{sec:formulation}). When survey data is collected with the intended purpose of downstream analysis (e.g. regression, classification), this formulation offers a more appropriate testing criterion to compare the credibility of survey data with respect to validation data. This provides a novel task-specific perspective compared to classical distribution testing problems. It is important to note that the proposed test is not a strictly weaker problem than identity testing as two different distributions can have the same optimal model, e.g. two far-apart distributions can yield the same linear model.

\noindent\textbf{2. A Generic Algorithm Design.} We develop a generic algorithmic framework \SurVerif{} to conduct credibility testing of survey data with bounded variables, and independent, homoscedastic noise (Section~\ref{Section : Algorithm for testing credibility}). 
Specifically, for both public and private surveys, \SurVerif{} accepts $\sampD$ to be credible with high probability if $\distF(f_\surveyset,\optimalf)$ is smaller than a threshold, and rejects only if $\distF(f_\surveyset,\optimalf)$ is larger than a threshold. 
Under the stated assumptions, we prove the correctness and sample complexity of \SurVerif{} for both public and $(\alpha, \beta)$-local DP surveys (Section~\ref{Section : Algorithm for testing credibility}). 

\noindent\textbf{3. Regression with Local DP.} Since \SurVerif{} needs to learn a regression model to test credibility, we need to derive bounds on the estimation error of linear regression models trained on private survey data. 
Hence, we propose a generalized technique, namely \AlgoPriv, to publish surveys satisfying LDP guarantees, and \AlgoReg{} to learn a Lasso regression on the privately published survey data $\dpsurveyset$ with high probability bounds on the estimation error (Section~\ref{sec:ldp_survey}). The proposed techniques also yield estimation error bounds for linear regression problems with subexponential covariates, and subexponential noise in the covariates and response. We prove that \AlgoReg{} achieves the optimal estimation error bound for $\ell_1$ linear regression~\citep{wang2019sparse}, which might be of broader interest.

Further details on related works are deferred to the Supplementary Material.

\section{Notations and Preliminaries}
\textbf{Notations.} We denote vectors with small bold letters (e.g. $\covariate$) and matrices with capital bold letters (e.g. $\covariatematrix$). 
The identity matrix of size $d$ is denoted by $\identity_d$.
For any matrix $\mathbf{M}$, we denote its minimum eigenvalue by $\lambda_{\min}(\mathbf{M})$.

\subsection{Linear Regression}\label{Section : Linear Regression}

Performing regression on survey data to fit reasonable models over the population is central to a wide variety of analysis tasks~\citep{Charvat2015_LR,Pan2017_LR,Meerwijk2017_LR}. Often, the observations collected to construct a survey data are the result of a complex sampling design reflecting the need to collect data as efficiently as possible within cost constraints. In the field of social science, it is comparatively unusual to find surveys that cover an area of any appreciable size which do not use stratified or multistage sampling~\citep{Groves2011survey,Lohr2021_survey-sampling,Kalton2020_survey-sampling}. Thus, verifying credibility of such survey data for correct downstream analysis is imperative.


Specifically, we use linear regression that tries to fit a linear model between the response and the covariates, i.e.
\begin{center}
    $\response = \innerproduct{\coeff}{\covariate} + \regnoise$\,.
\end{center}
Here, the response $\response$ depends on the covariates $\covariate$ via a vector of regression coefficients $\coeff\in \R^d$, while $\regnoise$ corresponds to the error representing the deviation of observations from the regression model. The underlying assumption in regression analysis is that a causal relationship exists between the response and covariates~\citep{della2023online}.

The linear regression model learned from a given survey sample $\surveyset$ is the $\coeff \in \R^d$ that best fits the survey data $\surveyset$, according to a pre-determined loss function. The most common loss function for linear regression analysis is the ordinary least squares loss $\sL(\response,\response') \triangleq (\response - \response')^2$.
Hence, we compute the linear regression model as $\coeff \triangleq \argmin_{\coeff \in \R^d} \sum_{(\covariate_i,\response_i) \in \surveyset} (\innerproduct{\coeff}{\covariate_i} - \response_i)^2$. 
In other words, solving the linear regression problem is to effectively pick out the best-fitting linear function (from the space of linear functions $\linfunc$) over the population learned from a given survey sample $\surveyset$ based on a given loss function. 



If the loss function $\Loss(\response,\response')$ chosen for the linear regression problem is bounded for all $\response,\response' \in \R$, it is known as the \textit{bounded linear regression problem}.
Additionally, Assumption~\ref{Assumption : Bounded values for response variable} also enforces us to only consider linear models with bounded coefficients. 



 \subsection{Local Differential Privacy}

 Following its success in US Census~\citep{Kenny2021_US-Census}, Differential Privacy (DP)~\citep{Dwork2006_DP-survey,Wasserman2010_DP-framework} has emerged as the gold standard to turn any survey data and downstream analysis private. definition ensures indistinguishability between neighboring datasets, defined as datasets that differ at exactly one point. We consider a stricter notion of differential privacy, known as local differential privacy~\citep{DBLP:conf/focs/KasiviswanathanLNRS08_local_dp_intro,DBLP:conf/focs/DuchiJW13_LocalPrivacyMinimax} where instead of datasets, we consider each data point individually. The notion of neighboring datasets being indistinguishable here translates to any two points being indistinguishable with respect to the output of a local DP-preserving mechanism. 
\begin{definition}[\textbf{$(\dpepsilon,\dpdelta)$-Local Differential Privacy}]\label{Definition: LDP}
    A function $\func{f}{\xdomain}{\yrange}$ is said to be $(\dpepsilon,\dpdelta)$-local differential private if for any $\covariate,\covariate'\in \xdomain$ and for all $S \subseteq \range(f)$, we have:
    $\Pr[f(\covariate) \in S] \leq \exp(\dpepsilon) \cdot \Pr[f(\covariate') \in S] + \dpdelta$.
\end{definition}

Note that setting $\dpdelta = 0$ recovers the $\dpepsilon$-local DP guarantees. 
Here, we focus on two additive mechanisms, namely Laplacian and Gaussian, to achieve DP that adds noise depending on the $\ell_p$-sensitivity of the corresponding function~\citep{Dwork2014_DP-book}.
\begin{definition}[\textbf{$\ell_p$-Sensitivity}]\label{Definition: LP Sensitivity}
    For a function $\func{f}{\xdomain}{\yrange}$, its $\ell_p$-sensitivity for any $p \geq 1$, denoted $\Delta_p(f)$ is defined as:
    $\Sensitivity_p(f) \triangleq \sup_{\covariate,\covariate' \in \xdomain} \norm{f(\covariate)-f(\covariate')}_{p}$.
\end{definition}

\begin{lemma}[\textbf{Gaussian Mechanism}]\label{Lemma: Gaussian Mechanism}
    For any $\dpepsilon >0$ and $\dpdelta\in (0,1]$, Gaussian mechanism $\sM$ ensuring $(\dpepsilon,\dpdelta)$-DP for any $\func{f}{\xdomain}{\R^d}$ with $\ell_2$-sensitivity $\Sensitivity_2(f)$ is $\sM(\covariate) \triangleq f(\covariate) + Z$, where $Z \sim \sN(0,\sigma^2\identity_d)$ and $\sigma = O\left(\frac{\Delta_2(f) \sqrt{\log{1/\dpdelta}}}{\dpepsilon}\right)$. 
\end{lemma}

\begin{lemma}[\textbf{Laplacian Mechanism}]\label{Lemma: Laplacian Mechanism}
    For any $\dpepsilon>0$, the Laplacian mechanism $\sM$ ensuring $\dpepsilon$-DP for any $\func{f}{\sX}{\R^d}$ with $\ell_1$-sensitivity $\Sensitivity_1(f)$ is $\sM(\covariate) \triangleq f(\covariate) + Z$, where $Z = \left(z_1,z_2,\ldots,z_d\right)$ and $z_i \sim \laplace \left(0,\frac{\Delta_1(f)}{\dpepsilon}\right)$. 
\end{lemma}

\section{\SurVerif: A Framework for Testing}\label{Section : Algorithm for testing credibility}


We now discuss the primary contribution of this paper: an efficient algorithmic framework \SurVerif{}, that verifies whether the linear regression model learned from a survey sample $\surveyset$ is not significantly far in statistical distance from the optimal model learned from a true distribution $\D$ by utilizing sampling access to $\D$. The pseudo-code of \SurVerif{} is presented in Algorithm \ref{alg:SurVerif}.

\SurVerif{} takes in as input a survey data $\surveyset\subset \mathbb{R}^{(d+1)}$, with  $|\surveyset| = \sizeS$, acceptance parameter $\K \in [0,1)$, rejection parameter $\tol \in (0,1]$, confidence parameter $\confidence \in (0,1]$, bound $\ybound$ for the response variable $\response$, weight bound $\linweight$ for the linear regression model to be learned and utilizes sampling access to the true distribution $\D$ and returns ACCEPT or REJECT with the following guarantees:

\begin{itemize}[leftmargin=*]
    \item If \SurVerif{} outputs REJECT, then with probability  $\geq (1 - \confidence)$ the given survey data $\surveyset$ is $(\kappa + \tol)$-far from being a credible data set with respect to $\blinfunc$.

    
    \item If the survey data $\surveyset$ is $\kappa$-close to being a credible data set with respect to $\blinfunc$ then \SurVerif{} outputs ACCEPT with probability at least $1 - \confidence$.
    
\end{itemize}

    

\setlength{\textfloatsep}{4pt}
\begin{algorithm}[h!]
\caption{\SurVerif($\surveyset \subset \mathbb{R}^{(d+1)},\D,\K,\confidence,\tol, \ybound, \linweight$)}\label{alg:SurVerif}
\begin{algorithmic}[1]
\State Initialize $\sampleD \gets \ceil*{\frac{\ybound^2\log (\frac{4}{\confidence})}{2\tol^2}}$, $\sizeS \gets |\surveyset|$, $\surveyset_{\D} \gets \emptyset$ \label{SurVerif Line 1}
\State $f_{\surveyset} \gets \argmin_{f \in \blinfunc} \frac{1}{\sizeS} \sum_{(\covariate,\response) \in \surveyset} (f(\covariate) - \response)^2$ \label{SurVerif Line 2}
\State $\hat{L} \gets \frac{1}{\sizeS}\sum_{(\covariate,\response) \in \surveyset} (f_{\surveyset}(\covariate) - \response)^2$ \label{SurVerif Line 3}
\State $\estimatevarS \gets \hat{L} + \frac{8\ybound\boundonx\linweight^2\sqrt{2\log{(2d)}}}{\sqrt{\sizeS}} +  3\ybound\sqrt{\frac{\log\frac{4}{\confidence}}{2\sizeS}}$ \label{SurVerif Line 4}
\State $\surveyset_{\D} \gets \sampleD \text{ iid samples from } \D$. \label{SurVerif Line 5}
\State $\estimatevarD \gets \frac{1}{\sampleD}\sum_{(\covariate,\response) \in \surveyset_{\D}} (f_\surveyset(\covariate) - \response)^2$ \label{SurVerif Line 6}
\If{$\sqrt{\estimatevarD} > \sqrt{\estimatevarS} + \K + \tol  $}~~Output REJECT. \label{SurVerif Line 7}
\Else ~~Output ACCEPT. \label{SurVerif Line 8}
\EndIf
\end{algorithmic}
\end{algorithm}

We provide a brief high-level overview of our proposed algorithmic framework \SurVerif{} before providing the main statement of the correctness of the algorithm. 




\paragraph{High-level Overview of the Correctness of \SurVerif.} The core idea of \SurVerif{} is that for evaluating the credibility of the survey data $\surveyset$ we proceed in two phases. In the first phase, (in lines \ref{SurVerif Line 2} to \ref{SurVerif Line 4}) we learn the linear regression model $f_{\surveyset}$ which fits the observations in $\surveyset$. In the second phase, (in lines \ref{SurVerif Line 5} and \ref{SurVerif Line 6}) we evaluate the credibility of $\surveyset$ by drawing sufficient samples from the true distribution $\D$ to obtain an additive estimate of the expected loss of the function $f_{\surveyset}$ for samples drawn from $\D$. Finally, (in Lines \ref{SurVerif Line 7} and \ref{SurVerif Line 8}) if this estimate differs significantly from the upper bound on the expected loss of $f_{\surveyset}$, \SurVerif{} REJECTS the survey data $\surveyset$ implying the survey set is far from being a credible survey data with respect to the class $\blinfunc$. We would also like to emphasize that by a careful treatment of generalization error bound with $\ell_1$ geometry, we obtain an upper bound on the expected loss on $\surveyset$, $\estimatevarS$, with $\sqrt{\log(d)}$ dependence rather than the classical $\sqrt{d}$ term~\citep[Theorem 11.8]{mohri2018}.

The correctness of \SurVerif{} follows from the following theorem the proof of which is included in the supplementary materials.
   




\begin{theorem}[\textbf{Correctness of \SurVerif}]\label{Theorem: Tester Accepts or Rejects}

Given a survey data $\surveyset$, sampling access to the true distribution $\D$,  along with parameters $\tol,\confidence \in (0,1]$ and $\boundony,\linweight,\K > 0$,

\begin{enumerate}[leftmargin=*]
    \item If the survey data $\surveyset$ is $\K$-close to a credible data set with respect to $\blinfunc$, then \emph{\SurVerif} outputs ACCEPT with probability $1 -\confidence$.

    \item However, if \emph{\SurVerif} outputs REJECT, then with probability at least $1 - \confidence$, the survey data $\surveyset$ is $\K$-far from being a credible data set with respect to $\blinfunc$.

    \item  As $\sizeS \rightarrow \infty$, if the survey data $\surveyset$ is $\K + 2\tol + 2\noisevariance$ far from a credible dataset with respect to $\blinfunc$, then \emph{\SurVerif} outputs REJECT with probability $1 - \confidence$.
\end{enumerate}
Also,  \emph{\SurVerif} requires at most $O(\frac{\boundony^2\log(\frac{4}{\confidence})}{\tol^2})$ samples from $\D$ for the validation set. 
\end{theorem}

\begin{remark}[Extensions to LDP Surveys]
\SurVerif{} can be suitably generalized to handle $\alpha$-LDP and \dppair-LDP survey data with similar sample complexity. Note that the threshold at which \SurVerif{} rejects a given survey data $\surveyset$ depends inversely on its size, i.e. $\sizeS$. Since there is no control over the size of the data $\surveyset$, \SurVerif{} does not reject unless it finds a reasonable certificate. $\alpha$-LDP survey data and \dppair-LDP survey data, the privacy cost is taken care of by the higher threshold set for the testing procedure.
\end{remark}

\begin{remark}[One-Sidedness of \SurVerif{}]
Note that \SurVerif{} is 1-sided as it ACCEPTS unless it finds a reasonable certification that the dataset $\surveyset$ is far from being credible with respect to $\blinfunc$. So it is possible that the dataset $\surveyset$ is not close to being credible with respect to $\blinfunc$ and yet  \SurVerif{} will ACCEPT. 
\end{remark}









\begin{remark}[Access to Misspecified $\valD$]
In Line~\ref{SurVerif Line 5} of \SurVerif{}, we are generating $\sampleD$ iid samples from the true distribution $\D$. We may not have sampling access to the true distribution $\D$ but access to a distribution $\sD'$ which is $\omega$-close to the true distribution in total variation distance. In that case, using Data Processing Inequality, we can guarantee credibility of the survey with high probability.
\end{remark}



\section{Publishing Surveys \& Regression with LDP}\label{sec:ldp_survey}
Now, we discuss our other contribution: a framework (\AlgoPriv) to publish survey data satisfying LDP guarantees, and an algorithm (\AlgoReg) to learn Lasso regression on the privately published survey data. 
Given survey data satisfying the bounded response variable and covariates assumption (Assumption~\ref{Assumption : Bounded values for response variable}), \AlgoPriv{} publishes private survey data satisfying LDP guarantees and the covariance of added noise. \AlgoReg{} takes the private survey data and noise covariance as input and outputs an estimate of the linear regression that is close to the true coefficients. The pseudocode of \AlgoPriv{} and \AlgoReg{} are presented in Algorithm~\ref{Algorithm: Gen LDP Survey} and~\ref{Algorithm: LDP Survey Regression}, respectively.

\subsection{Publishing Survey with LDP} 
\AlgoPriv{} takes as input a survey data $\surveyset$, the bound on each coordinate of the $\covariate$ as $\boundonx$, and the privacy parameters $\dpepsilon$ and $\dpdelta$. To ensure $\dpepsilon$-LDP, \AlgoPriv{} generates i.i.d. noise from a zero mean Laplace distribution with variance $\frac{8\boundonx^2}{\dpepsilon^2}$ and add that to each component of each $\covariate$ in $\surveyset$. To ensure \dppair-LDP, it similarly generates i.i.d. noise from a zero mean Gaussian with variance $\frac{8\boundonx^2}{\dpepsilon}\ln({\frac{1.25}{\dpdelta}})$ and proceeds similarly. Finally, it outputs the collection of the noisy covariates, i.e. $\dpsurveyset$, as the survey satisfying LDP guarantees and also the corresponding empirical covariance matrix $\noisevariance$ of the added Laplacian/Gaussian noise. 


\begin{algorithm}
    \caption{$\AlgoPriv\left(\surveyset,\dpepsilon,\dpdelta,\boundonx\right)$}\label{Algorithm: Gen LDP Survey}
    \begin{algorithmic}[1]
        \If{$\dpdelta \neq 0$}~~Set $\sD_\noisevariable = \sN\left(\mu = 0,\sigma^2 = \frac{\constant\boundonx}{\dpepsilon}\sqrt{\log\frac{1}{\dpdelta}}\right)$; 
        \Else~~Set $\sD_\noisevariable = \laplace\left(0,\frac{2\boundonx}{\dpepsilon}\right)$ \EndIf
        \State Initialize $\dpsurveyset \leftarrow \emptyset$ and $\noisevariance \leftarrow \Var(\noisedist(\dpepsilon,\dpdelta,\boundonx))\identity_d$\label{AlgoPriv Line 1}
        \For{$(\covariate,\response) \in \surveyset$}\label{AlgoPriv Line 2}
            \State Generate noise $\noisevariable$ using $\noboldnoisevariable_i \sim \noisedist(\dpepsilon,\dpdelta,\boundonx), \forall i \in [d]$\label{AlgoPriv Line 3}
            \State $\noisycovariate \leftarrow \noisevariable + \covariate$\label{AlgoPriv Line 4}
            \State $\dpsurveyset \leftarrow \dpsurveyset \cup (\noisycovariate,\response)$\label{AlgoPriv Line 5}
        \EndFor 
        \State\Return $(\dpsurveyset,\noisevariance)$\label{AlgoPriv Line 6}
    \end{algorithmic}
\end{algorithm}
Now, we show that \AlgoPriv{} ensures $\dpepsilon$- and \dppair-LDP for private survey data $\dpsurveyset$.

\begin{lemma}\label{lemma: Gaussian LDP Survey}
Given a survey data $\surveyset$, \emph{\AlgoPriv} yields 
\begin{enumerate}
\item \textbf{\dppair-Local DP Survey} If $\beta \neq 0$, \emph{\AlgoPriv} outputs $\dpsurveyset$ satisfying \dppair-local differential privacy and $\noisevariance = \frac{\constant\boundonx}{\dpepsilon}\sqrt{\log\frac{1}{\dpdelta}}\identity_d$ for some positive constant $\constant$.
\item \textbf{$\dpepsilon$-Local DP Survey} If $\beta = 0$, \emph{\AlgoPriv} outputs $\dpsurveyset$ satisfying $\dpepsilon$-local DP and $\noisevariance = \frac{8\boundonx^2}{\dpepsilon^2}\identity_d$.    
\end{enumerate}

\end{lemma}
    
\noindent\textit{Proof Sketch.} The results are direct consequences of Lemma~\ref{Lemma: Gaussian Mechanism} and~\ref{Lemma: Laplacian Mechanism} with the boundedness assumption (Assumption~\ref{Assumption : Bounded values for response variable}) guaranteeing that for a function $f(\covariate_i) = \covariate_i$ the $\ell_1$-sensitivity is $\Sensitivity_1(f) \leq 2\boundonx$.

\subsection{Regression on Noisy Covariates}
Now, we present a Lasso-based~\citep{tibshirani1996regression} regression algorithm, \AlgoReg{}, to perform linear regression with noisy covariates. Thus, we use it further to learn a linear model from LDP survey data.

\AlgoReg{} takes as input a noisy survey data $\dpsurveyset$, the corresponding noise covariance matrix $\noisevariance$, and a norm bound $R$ of the regression coefficients.
First, it uses the covariates of the noisy survey to compute the design matrix denoted by $\frac{1}{\sizeS}\sum_{i=1}^{\sizeS} \noisycovariate_i\noisycovariate_i^\top \triangleq \frac{1}{\sizeS}\noisycovariatematrix^T\noisycovariatematrix$, where $\noisycovariatematrix \in \R^{\sizeS \times d}$ is called the covariate matrix (or data matrix). Due to the noise in the covariates, we calibrate the design matrix further to compute the noisy design matrix $\basecovardummy \triangleq \frac{1}{\sizeS}\noisycovariatematrix^T\noisycovariatematrix - \noisevariance$ (Line \ref{AlgoReg Line 1}). \textit{This is the main deviation from the classical Lasso}. Then, we use the noise covariates and the response variable to compute $\predcovardummy \triangleq \frac{1}{\sizeS}\sum_{i=1}^{\sizeS} y_i \noisycovariate_i \triangleq \frac{1}{\sizeS}\noisycovariatematrix^T\responsevector$ (Line \ref{AlgoReg Line 2}).  $\basecovardummy$ and $\predcovardummy$ are the unbiased empirical estimates of $\xvariance$ and $\xvariance\coeff^*$, respectively. 
Now, we plug in these matrices in the Lasso optimization problem and further estimate the regression coefficients as
$\coeffhat \triangleq \argmin_{\norm{\coeff}_1 \leq \lassobound} \frac{1}{2}\coeff^T\basecovardummy\coeff - \inn{\predcovardummy}{\coeff}$. Note that though we present \AlgoReg{} in terms of LDP survey data, it works without modification for any setup with noisy covariates.


\begin{algorithm}
    \caption{$\AlgoReg(\dpsurveyset,\noisevariance, \lassobound)$}\label{Algorithm: LDP Survey Regression}
    \begin{algorithmic}[1]
        \State $\basecovardummy \leftarrow \frac{1}{\sizeS} \noisycovariatematrix^T\noisycovariatematrix - \noisevariance$\label{AlgoReg Line 1}
        \State $\predcovardummy \leftarrow \frac{1}{\sizeS} \noisycovariatematrix^T\responsevector$\label{AlgoReg Line 2}
        \State $\coeffhat \leftarrow \argmin_{\norm{\coeff}_1 \leq \lassobound} \frac{1}{2} \coeff^T\widehat{\Gamma}\coeff - \inn{\predcovardummy}{\coeff}$\label{AlgoReg Line 3}
        \State\Return $\coeffhat$\label{AlgoReg Line 4}
    \end{algorithmic}
\end{algorithm}



\noindent\textbf{Bounds on Estimation Error.} Assuming that the original survey data $\surveyset$ is generated from a linear model $\response = \inn{\coeff^*}{\covariate}$ with $\norm{\coeff^*}_1 \leq \lassobound$, we show that estimation error $\norm{\coeffhat - \coeff^*}_2$ of \AlgoReg{} on $\dpsurveyset$ is $O\left({\sizeS}^{-1/2}\right)$, where $n$ is sufficiently large. 

\begin{theorem}[\textbf{Learning From \dppair-Local DP Survey}]\label{Theorem: Learning from Gaussian LDP data}
    Let us consider a survey data $\surveyset$ with $\sizeS \geq  \max\left(\frac{
    \constant}{\lambda^2_{min}(\xvariance)}\left(\boundonx^2+\frac{\boundonx^2 \log{\left(\frac{1}{\dpdelta}\right)}}{\dpepsilon^2}\right)^2d\log{d}, 1\right)$ samples generated from a linear model $\response = \inn{\coeff}{\covariate} + \regnoise$ satisfying $\norm{\coeff^*}_1 \leq \lassobound$, and $|\covariatei_i|\leq \boundonx, \forall i \in [d]$, where $\regnoise$ comes from a subgaussian distribution with parameter $\sigma_\regnoise$. Now, if we apply \emph{\AlgoPriv} satisfying \dppair-local DP on $\surveyset$, then run \emph{\AlgoReg} to obtain $\coeffhat$. Then, for some constants $c_1$ and $c_2$, with probability at least $1-d^{-\constant_1}$, $\norm{\coeff^*-\coeffhat}_2$ is at most:
    \begin{align*}
    \small{
      \constant_2\frac{\boundonx\sqrt{\frac{\log\left(\frac{1}{\dpdelta}\right)}{\dpepsilon}+1}\left(\frac{\boundonx\sqrt{\log\left(\frac{1}{\dpdelta}\right)}}{\dpepsilon}+\sigma_\regnoise\right)}{\lambda_{\min}(\xvariance)}\norm{\coeff^*}_2\sqrt{\frac{d\log{d}}{\sizeS}}
    }
    \end{align*}
\end{theorem}
To prove, we use Lemma~\ref{lemma: Gaussian LDP Survey} for publishing the LDP survey with Gaussian mechanism and a result of~\cite{LDP_Regression_AOS} to bound estimation error in linear regression with any subgaussian additive noise on the covariates.
\begin{theorem}[\textbf{Learning from $\dpepsilon$-Local DP Survey}]\label{Theorem: Learning from Laplacian LDP data}
    Under the same premises of Theorem~\ref{Theorem: Learning from Gaussian LDP data}, let us consider that the noise $\regnoise$ to be generated from a sub-exponential distribution such that $\Pr[\regnoise \geq t] \leq \exp\left(-\frac{t}{\constant_\regnoise}\right)$, and the survey size 
\small{$\sizeS \geq \max\bigg( \max\left(\frac{\max\left(\frac{\boundonx}{\dpepsilon},\boundonx^2, \constant_\regnoise\right)}{\lambda_{\min}(\xvariance)},1\right)d\log{d}, 
 \max\left(\frac{\boundonx}{\dpepsilon},\boundonx^2, \constant_\regnoise\right) \log^3(d)\bigg)$}
Then, if we apply \emph{\AlgoReg} on an $\dpepsilon$-LDP version of $\surveyset$ published by \emph{\AlgoPriv} to get $\coeffhat$, for some constants $c_1, c_2 >0$, we obtain with probability at least $1-d^{-\constant_1}$, $\norm{\coeff^*-\coeffhat}_2$ is at most:
\begin{align*}
     \frac{\constant_2}{\lambda_{\min}(\xvariance)}\max\left(\frac{\boundonx}{\dpepsilon},\boundonx^2, \constant_\regnoise\right)\norm{\coeff^*}_2\sqrt{\frac{d\log{d}}{\sizeS}}\,
\end{align*}

\end{theorem}
To account for the subexponential noise in the $\covariate$ arising from \AlgoPriv{} with $\dpdelta = 0$, we prove the following result for \AlgoReg.

\begin{theorem}[\textbf{Learning from data with additive subexponential noise}]\label{Theorem: Sub-Expoential Noise Noisy Regression}
    Given a linear regression problem $\response = \inn{\coeff}{\covariate} + \regnoise, \covariate \in \R^d$ with optimal solution $\coeff^*$ with $\norm{\coeff^*}_1 \leq \lassobound$, where we observe $\noisycovariate = \covariate + \noisevariable$ and $\covariate$ comes from a distribution such that $\Pr[x_i \geq t] \leq \exp\left(-{t}/{\constant_\covariate}\right)$, $\regnoise$ comes from a distribution with $\Pr[\regnoise \geq t] \leq \exp\left(-\frac{t}{\constant_\regnoise}\right)$, and $\noisevariable$ comes from a distribution such that $\Pr[\noboldnoisevariable_i \geq t] \leq \exp\left(-\frac{t}{\constant_\noisevariable}\right)$, \emph{\AlgoReg} or its Lagrangian version with $\sizeS \geq \max\left( \max\{\frac{\constant_{\max}}{\lambda_{\min}(\xvariance)},1\}d\log{d}, \constant_{\max} \log^3(d)\right)$ satisfies
    \[
    \norm{\coeffhat-\coeff^*}_2 \leq \frac{\constant_1}{\lambda_{\min}(\xvariance)}\constant_{max}\norm{\coeff^*}_2\sqrt{\frac{d\log{d}}{\sizeS}}
    \]
    With probability at least $1 - d^{-\constant_2}$ for some constants $\constant_1$ and $\constant_2$, where $\constant_z = \constant_\covariate+\constant_\noisevariable$ and $\constant_{max} = \max\{\constant_\covariate,\constant_\noisevariable,\constant_\regnoise\}$.
\end{theorem}

\noindent\textit{Proof Sketch.} The proof is broadly divided into two parts. In the first part, we show that if $\norm{\predcovardummy - \basecovardummy\coeff^*}_\infty$ remains sufficiently small, then the estimation error $\norm{\coeff^* - \coeffhat}_2$ also remains small. 
We then need to prove that $\norm{\predcovardummy - \basecovardummy\coeff^*}_\infty$ indeed remains small when $\covariate$ and $\noisevariable$ comes from subexponential distributions. This requires us to prove a concentration result for subweibull distributions.


\begin{remark}[Subgaussian vs. subexponential noise] The bounds for estimation error for \AlgoReg{} in the case of $\dpepsilon$-LDP remains of the same order as that of \dppair-LDP. However, for subexponential noise, one requires a lower bound on the number of samples required to have the estimation error bound, i.e. $\sizeS \geq \max\left\{\constant_\regnoise^2,\frac{\boundonx^2}{\dpepsilon^2}\right\}\log^3(d)$. 
\end{remark}

\begin{remark}[Generic Analysis.] It is important to note that beyond the direct use case presented here for learning from $\dpepsilon$-LDP data, Theorem~\ref{Theorem: Sub-Expoential Noise Noisy Regression} ensures that \AlgoReg{} works for the general problem of learning regression when the $\covariate$, $\response$, $\regnoise$ and $\noisevariable$ are generated from subexponential distributions.
\end{remark}

\begin{remark}[Regularity of $\lambda_{\min}(\xvariance)$.] Note that the bound on estimation error depends on the smallest eigenvalue of the data covariance matrix $\xvariance$, i.e. $\lambda_{\min}(\xvariance)$, which is further dependent on the covariates. For example, if each coordinate of $\covariate$ is i.i.d., then the covariance matrix will be a diagonal matrix with the smallest non-zero element giving us the smallest eigenvalue. Furthermore, if the covariance matrix is full rank, then this value remains well-defined. However, for the case of non-full rank $\xvariance$, regularization techniques, like ridge and Lasso, can be used to lower bound $\lambda_{\min}(\xvariance)$, which in turn yields an upper bound on estimation error.
\end{remark}
\begin{remark}[Optimality]
    We observe that the estimation upper bound of \AlgoReg{} matches with the existing lower bound for sparse linear regression under LDP~\citep[Theorem 1]{wang2019sparse} up to logarithmic factors. The algorithm proposed by~\citep{wang2019sparse} requires exact knowledge of sparsity parameter and a Restricted Isometry Property (RIP) of the data to achieve the similar order of upper bound, whereas we do not need any such assumption and thus closes the gap with the existing lower bound. Additionally, we require $\max\{\frac{d\log d}{\min\{\alpha,1/\zeta\}}, \frac{d\log d}{\min\{\alpha,1/\zeta\}}\}$ initial samples to achieve this upper bound, whereas \citep{wang2019sparse} need $\frac{d\log d}{\alpha}$ samples. Thus, for $\alpha>1/\zeta$ (which includes most of the practical privacy levels), \AlgoReg{} is significantly sample efficient than existing sparse linear regression algorithms.
\end{remark}

\section{Experimental Analysis}
In this section, we numerically verify whether \SurVerif{} ACCEPTs or REJECTs  as per the theoretical analysis, 
and also the efficiency of \AlgoPriv{} and \AlgoReg{} to learn a linear regression model on LDP survey data. 

\paragraph{Experimental Setup.} We implement all the algorithms in Python 3.10. We use \texttt{LinearRegression} from \texttt{scikit-learn} to learn $\surveyf$. We run our simulations on Google Collaboratory with 2 Intel(R) Xeon(R) CPU @ 2.20GHz, 12.7GB RAM, and 107.7GB Disk Space.

\begin{figure*}[t!]
\centering
\begin{minipage}{0.48\textwidth}
\includegraphics[width=\linewidth]{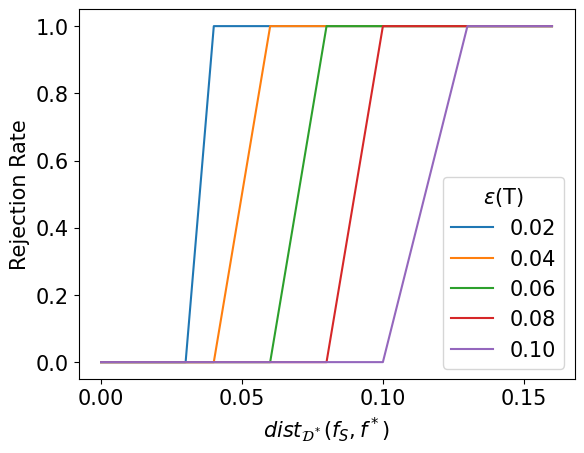}\vspace*{-.5em}
\caption{\small{Rejection rate of \SurVerif{} on \texttt{Synthetic\_1} vs. model distance (mean over 30 runs) for $\delta$ = 0.1 and different tolerance parameters $\epsilon$.}}\label{Fig: Model Distance}
\end{minipage}\hfill
\begin{minipage}{0.48\textwidth}
\includegraphics[width=\linewidth]{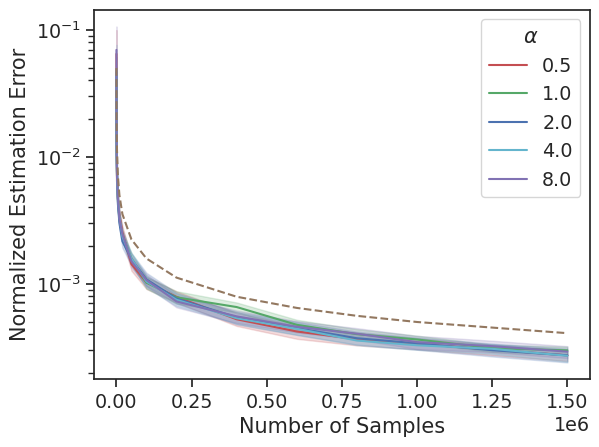}\vspace*{-.5em}
\caption{\small{Estimation error of \AlgoReg{} (mean $\pm$ std. over 30 runs) for \dppair-LDP version of \acsincome with $\dpdelta = 0.1$ and different $\dpepsilon$.}}\label{Fig: dppair LDP}
\end{minipage}\\
\begin{minipage}{0.48\textwidth}
\includegraphics[width=\linewidth]
{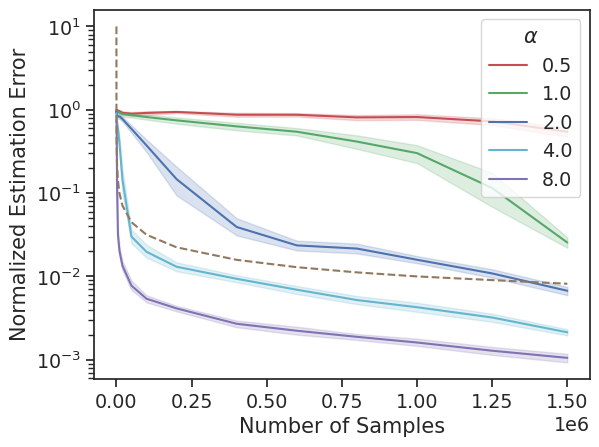}\vspace*{-.5em}
\caption{\small{Estimation error of \AlgoReg{} (mean $\pm$ std. over 30 runs) for $\dpepsilon$-LDP version of \acsincome for different $\dpepsilon$.\\}}\label{Fig: dpepsilon LDP}
\end{minipage}\hfill
\begin{minipage}{0.48\textwidth}
\includegraphics[width=\linewidth]{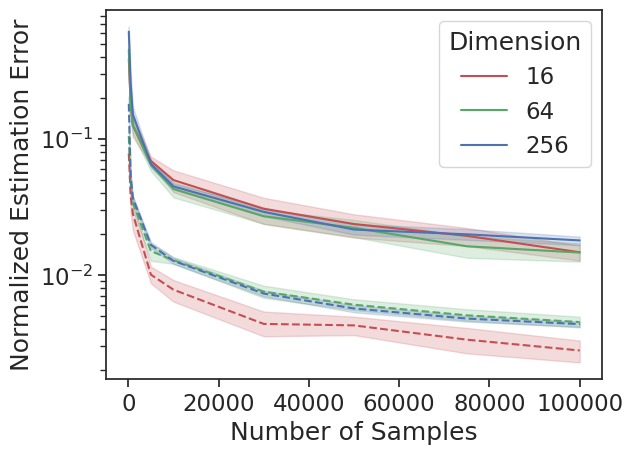}\vspace*{-.5em}
\caption{\small{Estimation error of \AlgoReg{} (mean $\pm$ std. over 30 runs) for subgaussian and subexponential noise for different dimensions of covariates.}}\label{Fig: Noisy Regression}
\end{minipage}
\end{figure*}

We use three setups for our experiments. 
\begin{enumerate}[leftmargin=*]
\item 
We generate a synthetic dataset, \texttt{Synthetic\_1}, where each coordinate of each $\covariate$ is generated from $\normal(0,1)$, and $\regnoise$ is generated from $\normal(0,0.1)$. For $\sampD$, we generate $\coeff_{\surveyset}$ such that each coordinate is generated from $\normal(0,0.01)$. For $\valD$, we generate the coefficients $\coeff^*$ with each coordinate being generated from $\normal(\mu,0.01)$ with $\mu$ taking values from $0$ to $2$ at intervals of $0.2$. As the value of $\mu$ increases the model distance between $f_\surveyset$ and $\optimalf$ increases.
\item  
We generate a synthetic dataset, \texttt{Synthetic\_2}, where each coordinate of $\covariate$ are generated from $\normal(0,1)$, and $\regnoise$ are generated from $\sN(0,1)$. The coefficients are generated by fixing values from $\uniform(1,10)$ with probability $\frac{1}{\sqrt{d}}$, and $0$ otherwise. We add noise with each coordinate, which are generated from $\normal(0,1)$ and $\laplace(0,\frac{1}{\sqrt{2}})$ for subgaussian and subexponential noise, respectively. This ensures that both the noises have the same variance. 
\item  \acsincome: For the real-world dataset, we consider the normalized \acsincome dataset, which exhibits well-known fairness issues between two sensitive groups `Male' and `Female'~\citep{DBLP:conf/nips/DingHMS21}. 
\end{enumerate}

\noindent\textbf{Objectives.} The objective of our experiments is to test:\\~\\
\textbf{Q1.}{ Does \SurVerif{} ACCEPT when the dataset $\surveyset$ is 
close to being a credible data set with respect to $\blinfunc$? Does \SurVerif{} REJECT when $\surveyset$ is 
far from any credible data set with respect to $\blinfunc$?\\~\\
\textbf{Q2.}{ How does the \AlgoReg{} perform on the survey data satisfying LDP?}\\~\\
\textbf{Q3.}{ How accurate is \AlgoReg{} under subgaussian and subexponential noise?}

\noindent\textbf{Experimental Results and Observations} are as follows:

\begin{table}[t!]
    \centering
    \caption{Performance of \SurVerif{} on \acsincome dataset with tolerance parameter $\epsilon = 0.1$.}\label{Table: Surverif ACS}
    \begin{tabular}{c|ccccccc}
    \toprule
     \#samples    & 100 & 1000 & 10000 & 50000 & 100000 & 600000 & 1000000 \\\midrule
     Rejection rate    & 0.57 & 0.60 & 0.70 & 0.83 & 0.97 & 1.00 & 1.00\\
     \bottomrule
    \end{tabular} 
\end{table}

\noindent\textbf{Q1: Performance of \SurVerif{} on Synthetic and Real-World Datasets.} In Figure~\ref{Fig: Model Distance}, we depict how \SurVerif{} performs with varying 
values of $\distF(\surveyf,\optimalf)$ under different values of tolerance parameter $\epsilon$ and confidence $\delta = 0.1$. In Table~\ref{Table: Surverif ACS}, we show how the rejection rate of \SurVerif{} changes with varying number of samples, where $\sampD$ and $\valD$ denoting the distribution of `Female' and `Male' in \acsincome, respectively. For \acsincome, we set the tolerance parameter to be $\epsilon=0.1$. 

\noindent\textbf{Observations.} (i) On the synthetic dataset \texttt{Synthetic\_1}, \SurVerif{} always rejects whenever the model distance crosses the specified tolerance parameter $\epsilon$. It also always accepts when the model distance is lower than the tolerance parameter $\epsilon$. (ii) On the \acsincome dataset, \SurVerif{} fails to estimate the error of $\surveyf$ on $\valD$ accurately for a small number of samples, while the accuracy of estimation improves with increasing number of samples.

\noindent\textbf{Q2: Performance of \AlgoPriv{} and \AlgoReg{} under \dppair- and $\dpepsilon$-LDP.} We implement \AlgoPriv{} and \AlgoReg{} on \acsincome for both $\dpepsilon$ and \dppair-LDP with $\dpepsilon \in \{0.5, 1.0, 2.0, 4.0, 8.0\}$. We fix $\dpdelta = 0.1$ throughout for \dppair-LDP. We evaluate the performance of these algorithms using different sample sizes from $100$ to $1500000$.
In Figure~\ref{Fig: dppair LDP} and~\ref{Fig: dpepsilon LDP}, we plot the change in Normalized Estimation Error, i.e. $\frac{\norm{\coeff^* - \coeffhat}_2}{\norm{\coeff^*}_2}$, with sample size for \dppair- and $\dpepsilon$-LDP, respectively. We also plot the theoretical convergence rates with dashed lines in the corresponding figures.


\noindent\textbf{Observations.} (i) \AlgoReg{} ensures that the Normalized Estimation Error for the coefficients decrease as per the specified rates of $O(1/\sqrt{\sizeS})$ and $O(1/{\sizeS})$ for both $\dpepsilon$ and \dppair-LDP data but only after crossing the initial threshold for number of samples. 
(ii) Normalized Estimation Error for \AlgoReg{} decays at a slower rate for $\dpepsilon$-LDP compared to the case for \dppair-LDP privacy. 
(iii) The initial threshold of sample complexity appears to be larger for $\dpepsilon$-LDP compared to the case for \dppair-LDP privacy. These results are reflective of the bounds obtained in Theorem~\ref{Theorem: Learning from Gaussian LDP data} and~\ref{Theorem: Learning from Laplacian LDP data}. 



\noindent\textbf{Q3: Performance of \AlgoReg{} under Subgaussian and Subexponential Noise.} To compare the performance of \AlgoReg{} for subgaussian and subexponential noise, we use the synthetic dataset \texttt{Synthetic\_2} consisting of Laplacian and Gaussian noise of same variance. We depict the results of our simulation in Figure~\ref{Fig: Noisy Regression} with the dashed lines and solid lines representing the cases for subgaussian and subexponential noises, respectively.

\noindent\textbf{Observations.} We observe that \AlgoReg{} performs better in case of subgaussian noise than subexponential noise as predicted by the theoretical results.


\section{Conclusion}
We propose an algorithm, \SurVerif, for testing the credibility of survey data in both public and public setup using linear regression models. \SurVerif{} does this by testing the closeness of models learned which we believe is first time in the testing literature. In the process, we propose \AlgoPriv{} to publish local DP survey and \AlgoReg{} to learn with high probability a linear regression model on a local DP survey. 
\AlgoReg{} also serves as a mechanism to learn linear regression models from data corrupted with noise coming from any subexponential distribution. We
prove that \AlgoReg{} achieves the optimal estimation error bound for $\ell_1$ linear regression in the LDP setting, which might be of broader interest. 
Finally, we numerically validate our theoretical results.

Here, we have specifically focused on linear regression as the task to test the credibility of the survey. One might like to extend this framework to other tasks, like kernel regression, principal component analysis, and others.
\bibliographystyle{apalike}
\bibliography{ref}

\begin{thebibliography}{}

\bibitem[Acharya et~al., 2014]{Acharya2014_sublinear}
Acharya, J., Jafarpour, A., Orlitsky, A., and Suresh, A.~T. (2014).
\newblock Sublinear algorithms for outlier detection and generalized closeness
  testing.
\newblock In {\em 2014 IEEE International Symposium on Information Theory},
  pages 3200--3204. IEEE.

\bibitem[Alabi et~al., 2022]{Diff_private_Simple_regression_AlabiMSSV22}
Alabi, D., McMillan, A., Sarathy, J., Smith, A.~D., and Vadhan, S.~P. (2022).
\newblock Differentially private simple linear regression.
\newblock {\em Proc. Priv. Enhancing Technol.}, 2022(2):184--204.

\bibitem[Amin et~al., 2023]{ICLR23_Diff_private_LinReg}
Amin, K., Joseph, M., Ribero, M., and Vassilvitskii, S. (2023).
\newblock Easy differentially private linear regression.
\newblock In {\em The Eleventh International Conference on Learning
  Representations, {ICLR} 2023}.

\bibitem[Awasthi et~al., 2020]{awasthi2020rademacher}
Awasthi, P., Frank, N., and Mohri, M. (2020).
\newblock On the rademacher complexity of linear hypothesis sets.

\bibitem[Bakhshizadeh et~al., 2023]{Sharp_SubWeibull_Bakhshizadeh}
Bakhshizadeh, M., Maleki, A., and de~la Pena, V.~H. (2023).
\newblock {Sharp concentration results for heavy-tailed distributions}.
\newblock {\em Information and Inference: A Journal of the IMA},
  12(3):1655--1685.

\bibitem[Balia and Jones, 2008]{Balia2008_survey}
Balia, S. and Jones, A.~M. (2008).
\newblock Mortality, lifestyle and socio-economic status.
\newblock {\em Journal of health economics}, 27(1):1--26.

\bibitem[Banerjee et~al., 2020]{banerjee2020governance}
Banerjee, A., Duflo, E., Imbert, C., Mathew, S., and Pande, R. (2020).
\newblock E-governance, accountability, and leakage in public programs:
  Experimental evidence from a financial management reform in india.
\newblock {\em American Economic Journal: Applied Economics}, 12(4):39--72.

\bibitem[Batu et~al., 2002]{Batu2022_DUAL}
Batu, T., Dasgupta, S., Kumar, R., and Rubinfeld, R. (2002).
\newblock The complexity of approximating entropy.
\newblock In {\em STOC2002}, pages 678--687.

\bibitem[Batu et~al., 2001]{Batu2001testing}
Batu, T., Fischer, E., Fortnow, L., Kumar, R., Rubinfeld, R., and White, P.
  (2001).
\newblock Testing random variables for independence and identity.
\newblock In {\em FOCS2001}, pages 442--451. IEEE.

\bibitem[Batu et~al., 2000]{Batu2000_equivalence}
Batu, T., Fortnow, L., Rubinfeld, R., Smith, W.~D., and White, P. (2000).
\newblock Testing that distributions are close.
\newblock In {\em FOCS2000}, pages 259--269. IEEE.

\bibitem[Bhattacharyya and Chakraborty, 2018]{Bhattacharyya2018_subcube-cond}
Bhattacharyya, R. and Chakraborty, S. (2018).
\newblock Property testing of joint distributions using conditional samples.
\newblock {\em ACM Transactions on Computation Theory (TOCT)}, 10(4):1--20.

\bibitem[Cai et~al., 2023]{cai2023scoreattack_lowerbound}
Cai, T.~T., Wang, Y., and Zhang, L. (2023).
\newblock Score attack: A lower bound technique for optimal differentially
  private learning.

\bibitem[Canonne et~al., 2014]{Canonne2014_equivalent}
Canonne, C., Ron, D., and Servedio, R.~A. (2014).
\newblock Testing equivalence between distributions using conditional samples.
\newblock In {\em Proceedings of the twenty-fifth annual ACM-SIAM symposium on
  Discrete algorithms}, pages 1174--1192. SIAM.

\bibitem[Canonne, 2015]{Canonne:Survey:ToC}
Canonne, C.~L. (2015).
\newblock A survey on distribution testing: Your data is big. but is it blue?
\newblock {\em Electron. Colloquium Comput. Complex.}, {TR15-063}.

\bibitem[Canonne, 2022]{CanonneTopicsDT2022}
Canonne, C.~L. (2022).
\newblock {Topics and Techniques in Distribution Testing: A Biased but
  Representative Sample}.
\newblock {\em {Foundations and Trends in Communications and Information
  Theory}}, 19(6):1032--1198.

\bibitem[Canonne et~al., 2021]{Canonne2021_subcube-cond}
Canonne, C.~L., Chen, X., Kamath, G., Levi, A., and Waingarten, E. (2021).
\newblock Random restrictions of high dimensional distributions and uniformity
  testing with subcube conditioning.
\newblock In {\em Proceedings of the 2021 ACM-SIAM Symposium on Discrete
  Algorithms (SODA)}, pages 321--336. SIAM.

\bibitem[Canonne et~al., 2015]{Canonne2015_cond}
Canonne, C.~L., Ron, D., and Servedio, R.~A. (2015).
\newblock Testing probability distributions using conditional samples.
\newblock {\em SIAM Journal on Computing}, 44(3):540--616.

\bibitem[Chakraborty et~al., 2013]{CFGM2013_conditional}
Chakraborty, S., Fischer, E., Goldhirsh, Y., and Matsliah, A. (2013).
\newblock On the power of conditional samples in distribution testing.
\newblock In {\em Proceedings of the 4th conference on Innovations in
  Theoretical Computer Science}, pages 561--580.

\bibitem[Chan et~al., 2014]{Chan2014_equivalence}
Chan, S.-O., Diakonikolas, I., Valiant, P., and Valiant, G. (2014).
\newblock Optimal algorithms for testing closeness of discrete distributions.
\newblock In {\em Proceedings of the twenty-fifth annual ACM-SIAM symposium on
  Discrete algorithms}, pages 1193--1203. SIAM.

\bibitem[Charvat et~al., 2015]{Charvat2015_LR}
Charvat, H., Goto, A., Goto, M., Inoue, M., Heianza, Y., Arase, Y., Sone, H.,
  Nakagami, T., Song, X., Qiao, Q., et~al. (2015).
\newblock Impact of population aging on trends in diabetes prevalence: a
  meta-regression analysis of 160,000 japanese adults.
\newblock {\em Journal of diabetes investigation}, 6(5):533--542.

\bibitem[Connors et~al., 2019]{Connors_Survey}
Connors, E.~C., Krupnikov, Y., and Ryan, J.~B. (2019).
\newblock {How Transparency Affects Survey Responses}.
\newblock {\em Public Opinion Quarterly}, 83(S1):185--209.

\bibitem[Dandekar et~al., 2018]{dandekar2018differential}
Dandekar, A., Basu, D., and Bressan, S. (2018).
\newblock Differential privacy for regularised linear regression.
\newblock In {\em International Conference on Database and Expert Systems
  Applications}, pages 483--491. Springer.

\bibitem[Daskalakis et~al., 2018]{Daskalakis2018_equivalence}
Daskalakis, C., Kamath, G., and Wright, J. (2018).
\newblock Which distribution distances are sublinearly testable?
\newblock In {\em Proceedings of the Twenty-Ninth Annual ACM-SIAM Symposium on
  Discrete Algorithms}, pages 2747--2764. SIAM.

\bibitem[Della~Vecchia and Basu, 2023]{della2023online}
Della~Vecchia, R. and Basu, D. (2023).
\newblock Online instrumental variable regression: Regret analysis and bandit
  feedback.
\newblock {\em arXiv preprint arXiv:2302.09357}.

\bibitem[Diakonikolas et~al., 2021]{Diakonikolas2021_equivalence}
Diakonikolas, I., Gouleakis, T., Kane, D.~M., Peebles, J., and Price, E.
  (2021).
\newblock Optimal testing of discrete distributions with high probability.
\newblock In {\em Proceedings of the 53rd Annual ACM SIGACT Symposium on Theory
  of Computing}, pages 542--555.

\bibitem[Ding et~al., 2017]{Ding2017_LDP}
Ding, B., Kulkarni, J., and Yekhanin, S. (2017).
\newblock Collecting telemetry data privately.
\newblock {\em Advances in Neural Information Processing Systems}, 30.

\bibitem[Ding et~al., 2021]{DBLP:conf/nips/DingHMS21}
Ding, F., Hardt, M., Miller, J., and Schmidt, L. (2021).
\newblock Retiring adult: New datasets for fair machine learning.
\newblock In Ranzato, M., Beygelzimer, A., Dauphin, Y.~N., Liang, P., and
  Vaughan, J.~W., editors, {\em NuerPS}, pages 6478--6490.

\bibitem[Dinur and Nissim, 2003]{DBLP:conf/pods/DinurN03}
Dinur, I. and Nissim, K. (2003).
\newblock Revealing information while preserving privacy.
\newblock In Neven, F., Beeri, C., and Milo, T., editors, {\em PODS}, pages
  202--210. {ACM}.

\bibitem[Duchi et~al., 2013]{DBLP:conf/focs/DuchiJW13_LocalPrivacyMinimax}
Duchi, J.~C., Jordan, M.~I., and Wainwright, M.~J. (2013).
\newblock Local privacy and statistical minimax rates.
\newblock In {\em FOCS}, pages 429--438. {IEEE} Computer Society.

\bibitem[Dwork, 2006]{Dwork2006_DP-survey}
Dwork, C. (2006).
\newblock Differential privacy.
\newblock In {\em International colloquium on automata, languages, and
  programming}, pages 1--12. Springer.

\bibitem[Dwork et~al., 2006]{Dwork2006_DP}
Dwork, C., Kenthapadi, K., McSherry, F., Mironov, I., and Naor, M. (2006).
\newblock Our data, ourselves: Privacy via distributed noise generation.
\newblock In {\em EuroCrypt}, pages 486--503. Springer.

\bibitem[Dwork and Roth, 2014]{Dwork2014_DP-book}
Dwork, C. and Roth, A. (2014).
\newblock The algorithmic foundations of differential privacy.
\newblock {\em Found. Trends Theor. Comput. Sci.}, 9(3–4):211–407.

\bibitem[Erlingsson et~al., 2014]{Erlingsson2014_rappor}
Erlingsson, {\'U}., Pihur, V., and Korolova, A. (2014).
\newblock Rappor: Randomized aggregatable privacy-preserving ordinal response.
\newblock In {\em Proceedings of the 2014 ACM SIGSAC conference on computer and
  communications security}, pages 1054--1067.

\bibitem[Evans et~al., 2022]{evans2022differentially}
Evans, G., King, G., Smith, A.~D., Thakurta, A., Katz, J., King, G.,
  Rosenblatt, E., Evans, G., King, G., Schwenzfeier, M., et~al. (2022).
\newblock Differentially private survey research.
\newblock {\em American Journal of Political Science}, 28:1--22.

\bibitem[Fisher, 2019]{Fisher2019_Fb}
Fisher, C. (2019).
\newblock Over 267 million facebook users reportedly had data exposed online.

\bibitem[Gibbs and Su, 2002]{Gibbs2002_metrics}
Gibbs, A.~L. and Su, F.~E. (2002).
\newblock On choosing and bounding probability metrics.
\newblock {\em International statistical review}, 70(3):419--435.

\bibitem[Government~of Canada, 2024]{statscanada}
Government~of Canada, S.~C. (2024).
\newblock Learning resources: Statistics: Power from data! non-probability
  sampling.
\newblock {\em \small{\texttt{https://www150.statcan.gc.ca/n1/edu
  /power-pouvoir/ch13/nonprob/5214898-eng.htm}}}.

\bibitem[Groves et~al., 2011]{Groves2011survey}
Groves, R.~M., Fowler~Jr, F.~J., Couper, M.~P., Lepkowski, J.~M., Singer, E.,
  and Tourangeau, R. (2011).
\newblock {\em Survey methodology}.
\newblock John Wiley \& Sons.

\bibitem[Heeringa et~al., 2017]{heeringa2017applied}
Heeringa, S.~G., West, B.~T., and Berglund, P.~A. (2017).
\newblock {\em Applied survey data analysis}.
\newblock chapman and hall/CRC.

\bibitem[Henriksen-Bulmer and Jeary, 2016]{Henriksen2016rei}
Henriksen-Bulmer, J. and Jeary, S. (2016).
\newblock Re-identification attacks—a systematic literature review.
\newblock {\em International Journal of Information Management},
  36(6):1184--1192.

\bibitem[Iyengar et~al., 2019]{iyengar2019towards}
Iyengar, R., Near, J.~P., Song, D., Thakkar, O., Thakurta, A., and Wang, L.
  (2019).
\newblock Towards practical differentially private convex optimization.
\newblock In {\em 2019 IEEE symposium on security and privacy (SP)}, pages
  299--316. IEEE.

\bibitem[Kalton, 2020]{Kalton2020_survey-sampling}
Kalton, G. (2020).
\newblock {\em Introduction to survey sampling}.
\newblock Number~35. Sage Publications.

\bibitem[Karmakar and Basu, 2024]{karmakar2024marich}
Karmakar, P. and Basu, D. (2024).
\newblock Marich: A query-efficient distributionally equivalent model
  extraction attack.
\newblock {\em Advances in Neural Information Processing Systems}, 36.

\bibitem[Kasiviswanathan et~al., 2011]{Kasiviswanathan2011_LDP}
Kasiviswanathan, S.~P., Lee, H.~K., Nissim, K., Raskhodnikova, S., and Smith,
  A. (2011).
\newblock What can we learn privately?
\newblock {\em SIAM Journal on Computing}, 40(3):793--826.

\bibitem[Kasiviswanathan et~al.,
  2008]{DBLP:conf/focs/KasiviswanathanLNRS08_local_dp_intro}
Kasiviswanathan, S.~P., Lee, H.~K., Nissim, K., Raskhodnikova, S., and Smith,
  A.~D. (2008).
\newblock What can we learn privately?
\newblock In {\em FOCS}, pages 531--540.

\bibitem[Kenny et~al., 2021]{Kenny2021_US-Census}
Kenny, C.~T., Kuriwaki, S., McCartan, C., Rosenman, E.~T., Simko, T., and Imai,
  K. (2021).
\newblock The use of differential privacy for census data and its impact on
  redistricting: The case of the 2020 us census.
\newblock {\em Science advances}, 7(41):eabk3283.

\bibitem[Loh and Wainwright, 2012]{LDP_Regression_AOS}
Loh, P.-L. and Wainwright, M.~J. (2012).
\newblock {High-dimensional regression with noisy and missing data: Provable
  guarantees with nonconvexity}.
\newblock {\em The Annals of Statistics}, 40(3):1637 -- 1664.

\bibitem[Lohr, 2021]{Lohr2021_survey-sampling}
Lohr, S.~L. (2021).
\newblock {\em Sampling: design and analysis}.
\newblock Chapman and Hall/CRC.

\bibitem[Meerwijk and Sevelius, 2017]{Meerwijk2017_LR}
Meerwijk, E.~L. and Sevelius, J.~M. (2017).
\newblock Transgender population size in the united states: a meta-regression
  of population-based probability samples.
\newblock {\em American journal of public health}, 107(2):e1--e8.

\bibitem[Mohri et~al., 2018]{mohri2018}
Mohri, M., Rostamizadeh, A., and Talwalkar, A. (2018).
\newblock {\em Foundations of machine learning}.
\newblock MIT Press.

\bibitem[Onak and Sun, 2018]{Onak-Sun2018_PR}
Onak, K. and Sun, X. (2018).
\newblock Probability--revealing samples.
\newblock In {\em International Conference on Artificial Intelligence and
  Statistics}. PMLR.

\bibitem[Ouhamma et~al., 2021]{ouhamma2021stochastic}
Ouhamma, R., Maillard, O.-A., and Perchet, V. (2021).
\newblock Stochastic online linear regression: the forward algorithm to replace
  ridge.
\newblock {\em Advances in Neural Information Processing Systems},
  34:24430--24441.

\bibitem[Pan, 2017]{Pan2017_LR}
Pan, W.-T. (2017).
\newblock A newer equal part linear regression model: A case study of the
  influence of educational input on gross national income.
\newblock {\em Eurasia Journal of Mathematics, Science and Technology
  Education}, 13(8):5765--5773.

\bibitem[Paninski, 2008]{Paninski2008_identity}
Paninski, L. (2008).
\newblock A coincidence-based test for uniformity given very sparsely sampled
  discrete data.
\newblock {\em IEEE Transactions on Information Theory}, 54(10):4750--4755.

\bibitem[Plutzer, 2019]{Plutzer_Survey}
Plutzer, E. (2019).
\newblock {Privacy, Sensitive Questions, and Informed Consent: Their Impacts on
  Total Survey Error, and the Future of Survey Research}.
\newblock {\em Public Opinion Quarterly}, 83(S1):169--184.

\bibitem[Salant and Dillman, 1994]{salant1994conduct}
Salant, P. and Dillman, D.~A. (1994).
\newblock How to conduct your own survey.
\newblock {\em Willey}.

\bibitem[Stanley et~al., 2008]{stanley2008meta}
Stanley, T.~D., Doucouliagos, C., and Jarrell, S.~B. (2008).
\newblock Meta-regression analysis as the socio-economics of economics
  research.
\newblock {\em The Journal of Socio-Economics}, 37(1):276--292.

\bibitem[Tang et~al., 2017]{Tang2017_LDP}
Tang, J., Korolova, A., Bai, X., Wang, X., and Wang, X. (2017).
\newblock Privacy loss in apple's implementation of differential privacy on
  macos 10.12.
\newblock {\em arXiv preprint arXiv:1709.02753}.

\bibitem[Tibshirani, 1996]{tibshirani1996regression}
Tibshirani, R. (1996).
\newblock Regression shrinkage and selection via the lasso.
\newblock {\em Journal of the Royal Statistical Society Series B: Statistical
  Methodology}, 58(1):267--288.

\bibitem[Valiant and Valiant, 2017]{Valiant2017_identity}
Valiant, G. and Valiant, P. (2017).
\newblock An automatic inequality prover and instance optimal identity testing.
\newblock {\em SIAM Journal on Computing}, 46(1):429--455.

\bibitem[Valiant, 2008]{Valiant2008_equivalence}
Valiant, P. (2008).
\newblock Testing symmetric properties of distributions.
\newblock In {\em Proceedings of the fortieth annual ACM symposium on Theory of
  computing}, pages 383--392.

\bibitem[Victor et~al., 2020]{Victor2020_Google}
Victor, D., Frenkel, S., and Kershner, I. (2020).
\newblock Personal data of all 6.5 million israeli voters is exposed.
\newblock {\em The New York Times. https://www. nytimes.
  com/2020/02/10/world/middleeast/israeli-voters-leak. html}.

\bibitem[Wainwright, 2019]{Wainwright_2019_HDS}
Wainwright, M.~J. (2019).
\newblock {\em High-Dimensional Statistics: A Non-Asymptotic Viewpoint}.
\newblock Cambridge Series in Statistical and Probabilistic Mathematics.
  Cambridge University Press.

\bibitem[Wang and Xu, 2019]{wang2019sparse}
Wang, D. and Xu, J. (2019).
\newblock On sparse linear regression in the local differential privacy model.
\newblock In {\em International Conference on Machine Learning}, pages
  6628--6637. PMLR.

\bibitem[Wasserman and Zhou, 2010]{Wasserman2010_DP-framework}
Wasserman, L. and Zhou, S. (2010).
\newblock A statistical framework for differential privacy.
\newblock {\em Journal of the American Statistical Association},
  105(489):375--389.

\bibitem[Wood et~al., 2018]{Wood2018differential}
Wood, A., Altman, M., Bembenek, A., Bun, M., Gaboardi, M., Honaker, J., Nissim,
  K., O'Brien, D.~R., Steinke, T., and Vadhan, S. (2018).
\newblock Differential privacy: A primer for a non-technical audience.
\newblock {\em Vand. J. Ent. \& Tech. L.}, 21:209.

\bibitem[Yang et~al., 2023]{MengmengYang2020}
Yang, M., Guo, T., Zhu, T., Tjuawinata, I., Zhao, J., and Lam, K.-Y. (2023).
\newblock Local differential privacy and its applications: A comprehensive
  survey.
\newblock {\em Computer Standards \& Interfaces}, page 103827.

\end{thebibliography}
\clearpage
\newpage
\appendix
\part{Appendix}
\parttoc
\clearpage
\section{Related Work}\label{Sec: Related Work}
Our line of work falls under the head of applications of property testing in analysis of survey data.

\subsection{Testing Closeness of Distributions}

Testing identity between a known and an unknown distribution was first introduced by \cite{Batu2001testing}. Since then, this problem has been widely studied for structured distributions with several results providing tighter bounds for sample complexity ~\citep{Paninski2008_identity,Valiant2017_identity}. The harder problem of testing equivalence between unknown distributions was first studied by \cite{Batu2000_equivalence} and optimal upper and lower have been given by a series of works done in this field ~\citep{Valiant2008_equivalence,Chan2014_equivalence,Daskalakis2018_equivalence,Diakonikolas2021_equivalence,Acharya2014_sublinear}. A long series of works in the field of distribution testing have proposed a hierarchy of sampling models ~\citep{Batu2022_DUAL,Onak-Sun2018_PR} including the \textit{conditional sampling model}  ~\citep{CFGM2013_conditional,Canonne2015_cond} and its variant of \textit{subcube conditioning} ~\citep{Bhattacharyya2018_subcube-cond,Canonne2021_subcube-cond}  which have greatly reduced the query complexity required to solve the problems of testing identity and equivalence for large distributions ~\citep{Bhattacharyya2018_subcube-cond,Canonne2014_equivalent}. These algorithms have greatly improved on the classical sampling model by crucially leveraging the power to make adaptive conditional queries to the distributions. However, even in the presence of such powerful query access models, testing identity and/or equivalence between two very high-dimensional distributions is a very expensive proposition. In the context of surveys done for learning a model over the characteristics of the population, it is not quite clear when and how to use the power of these sampling models. On the other hand, the sample complexity incurs an exponential blowup with respect to the dimension of the distribution in the classical sampling model, which renders the model inefficient in practice.


\subsection{Learning-based Tasks for Data Satisfying DP}

In this modern age of data-driven analysis, a significant amount of data has been generated and collected for various purposes including decision making and service improvement. This data can be acquired from end-user devices, which includes the private data of individuals and hence are highly private. Collecting and analyzing data from end-user devices like mobile phones has incurred serious privacy issues since such data contain various sensitive information pertaining to the users~\citep{cai2023scoreattack_lowerbound,karmakar2024marich}. What's even worse is that advanced data fusion and analysis techniques can be used to infer the daily habits and behaviour profiles of a large number of individuals, thus breaching their privacy.

Differential privacy~\citep{Dwork2006_DP,Dwork2014_DP-book} has been the existing de facto standard for preserving individual privacy and has been applied in a large number of applications.  Traditional differential privacy, also known as centralized-DP is typically realised by having a data curator to collect the user’s original data first and then releasing the noisy statistical information to the public. The data aggregator is assumed to be the most trusted one in this centralized model. However, even big reputable companies like Google have failed to safeguard their customer's privacy~\citep{Victor2020_Google}. For instance, it has been reported in 2018 that thousands of users had their private data on Google+ social network leaked by Google. In 2019, millions of users on Facebook had their private data including user IDs, phone numbers and names exposed online~\citep{Fisher2019_Fb}. All these instances of data privacy leakage have prompted the research community to work towards a more stringent model of privacy, doing away with the trust on third party data-handling institutions. 

In order to circumvent the trust issues associated with this third-party database manager, the idea of local differential privacy was proposed~\citep{Kasiviswanathan2011_LDP}. This technique involves data perturbation at the user level, which is then collected by a central server for further data analysis. LDP has received significant attention in the modern age and has been deployed by a large number of big companies~\citep{Erlingsson2014_rappor,Tang2017_LDP,Ding2017_LDP} to preserve data privacy. However, addition of too much noise to the entire dataset to ensure LDP limits the scope of its application towards downstream analysis tasks. 


\section{Notations}

\begin{table}[H]
    \centering
    \caption{Summary of notations.}\label{tab:notations}
    \begin{tabular}{c|c}
    \toprule
    Symbols & Definitions\\
    \midrule
    $\yrange$ & Range of Response Variable\\
$\xdomain$ & Range of Covariates\\
$\linfunc$ & Space of Linear Functions\\
$\blinfunc$ & Space of Linear Functions with Bounded Coefficients\\
$\surveyset$ & Survey Data\\
$\sampD$ & Sampling Distribution of Survey Data\\
$\valD$ & True Distribution\\
$\boundonx$ & Bound on each Dimensions of Covariates\\
$\boundony$ & Bound on Response Variables\\
$\lambda_{min}(\mathbf{M})$ & Minimum Eigenvalue of Matrix $\mathbf{M}$\\
$\norm{\mathbf{M}}_{p,q}$ & $\norm{\norm{\mathbf{M}_1}_p,...,\norm{\mathbf{M}_m}_p}_q$, where $\mathbf{M}_i$ is the $i$-th column of the matrix $\mathbf{M}$\\
$\norm{\mathbf{M}}_{max}$ & Element-wise Maximum Element of Matrix $\mathbf{M}$\\
\bottomrule
\end{tabular}
\end{table}

\section{Proofs of Section 3: Testing Credibility}
For the correctness of our algorithm, we need a lower bound on the $\boundony$.

\begin{lemma}\label{lem:xyW}
    Under \textbf{Assumption 1} and \textbf{Assumption 2}, as long as $\linweight \leq \frac{\boundony}{\boundonx}$
    for any $\covariate$, $$\surveyf(\covariate) \in [-\boundony,\boundony].$$
\end{lemma}

\begin{proof}
    $|\surveyf(\covariate)| = |\inn{\coeff}{\covariate}| \leq \|\coeff\|_1\|\covariate\|_\infty \leq \linweight\boundonx$.
\end{proof}

\noindent
We use the framework of $\ell$-Lipschitz loss function defined below. 

\begin{definition}\label{Definition: Lipschitz continuous loss function}
    \textbf{($\lip$-Lipschitz Loss Function).} 
    Given $\ell > 0$, a 
    loss function $\Loss \in \sF_{loss}$ is called $\lip$-Lipschitz if for any fixed $y \in \yrange \subseteq \R$ and $\covariate_1,  \covariate_2 \in \xdomain$, the loss function $\sL: \yrange \times \yrange \rightarrow \R$ satisfies 
    $|\sL(f(\covariate_1),y) - \sL(f(\covariate_2),y)| \leq \lip|f(\covariate_1) - f(\covariate_2)|$. 
\end{definition}


We get the following theorem for $\lip$-Lipschitz loss functions and bounded linear hypothesis classes as direct consequence of~\cite[Theorem~11.3]{mohri2018}.

\begin{theorem}\label{Theorem: Lipschitz Regression Generalization}
    Let $\Loss: \yrange \times \yrange \rightarrow \R$ be the least squares loss having Lipschitz constant $\lip$ and the range of our hypothesis $f_{\surveyset} \in \blinfunc$ be upper bounded by $\boundony > 0$. Then we have with probability $1 - \confidence$:
    \begin{align*}
        \E_{(\covariate,\response) \sim \sampD} \left[ ( {f_{\surveyset}(\covariate)-\response})^2 \right]&\leq \frac{1}{\sizeS}\sum_{i \in [\sizeS]} (f_{\surveyset}(\covariate_i)-\response_i)^2 + 2\lip\widehat{\sR_S}(\blinfunc) + 3\ybound\sqrt{\frac{\log\frac{2}{\confidence}}{2\sizeS}}
    \end{align*}
\end{theorem}


\cite{awasthi2020rademacher} have proved the following upper bound that matches upto constant factors the existing lower bound for empirical Rademacher complexity for bounded linear hypothesis classes.

\begin{theorem}\label{Theorem : Upper bounds for Rademacher complexity by Awasthi}
    Let $\blinfunc = \{\covariate \rightarrow \inn{\bw}{\covariate}:\norm{\bw}_2 \leq \linweight\}$ be a family of linear functions defined over $\R^d$ with bounded weight in $\ell_1$-norm. Then, the empirical Rademacher complexity of $\blinfunc$ for a sample $\surveyset = \{\covariate_1,\covariate_2,...,\covariate_{\sizeS}\}$ gives:
    \begin{align*}
        \widehat{\sR_S}(\blinfunc) \leq \frac{\linweight}{\sizeS}\norm{\bf\covariatematrix^T}_{1,\infty}\sqrt{2\log{(2d)}}
    \end{align*}
    Where $\bfX$ is the $d\times \sizeS$ matrix with explanatory variable $\covariate_i$ as the $i^{\text{th}}$ column, i.e. $\bf\covariatematrix = [\covariate_1,\covariate_2,..,\covariate_m]$.
\end{theorem}

Using Theorem~\ref{Theorem : Upper bounds for Rademacher complexity by Awasthi} and Theorem~\ref{Theorem: Lipschitz Regression Generalization} for bounded linear regression problem, we obtain the following upper bound for the expected loss.

\begin{lemma}\label{Lemma: Rad-Comp-free generalization bound}
    Let $\Loss: \yrange \times \yrange \rightarrow \R$ be the $\lip$-Lipschitz ordinary least squares loss function such that $\yrange = [-\ybound,\ybound]$ and that the underlying hypothesis $f \in \blinfunc$ is linear with bounded $\ell_1$ norm $\linweight$. Then we have with probability $1 - \confidence$:
    \begin{align*}
        \E_{(\covariate,\response) \sim \sampD}[(f_{\surveyset}(\covariate)-\response)^2] \leq &\frac{1}{\sizeS}\sum_{i \in [\sizeS]} (f_{\surveyset}(\covariate_i)-\response_i)^2 + 2\ell\frac{\boundonx\linweight\sqrt{2\log{(2d)}}}{\sqrt{\sizeS}} + 3\ybound\sqrt{\frac{\log\frac{2}{\confidence}}{2\sizeS}} 
    \end{align*}
\end{lemma}
    \begin{proof}
        We need to obtain a closed-form expression for the $(1,\infty)$-norm of the $ \sizeS \times d$ matrix $\covariatematrix^T$ in order to use the result of Theorem \ref{Theorem : Upper bounds for Rademacher complexity by Awasthi}. Since by assumption \ref{Assumption : Bounded values for response variable}, we have $\|\covariatematrix^T_i\|_2 \leq \boundonx\sqrt{m}$ where $\covariatematrix^T_1$ denotes the $i$-th column of the matrix $\covariatematrix^T$, we can upper bound $\|\bf\covariatematrix\|_{1,\infty}$ as follows

        \begin{align*}
            \norm{\covariatematrix}_{1,\infty} = \left\|\|\covariate_j\|_2\right\|_\infty \leq \boundonx
            \sqrt{m}
        \end{align*}

        Now, we can plug in this upper bound for the Rademacher complexity term in Theorem \ref{Theorem: Lipschitz Regression Generalization} to obtain the bound stated in this lemma.
    \end{proof}

\begin{lemma}\label{Lemma: Separation in $l_2$ distance}
    Let the data be originally generated from a distribution $\sampD$ over $\xdomain \times \yrange$. 

If $\distF(f_{\surveyset},\optimalf) \leq \K$, we have:
    \begin{align*}
        \sqrt{\E_{(\covariate,\response) \sim \valD} [(\surveyf(\covariate)-\response)^2]} \leq \K + \sqrt{\E_{(\covariate,\response) \sim \D} [(\optimalf(\covariate)-\response)^2]} 
    \end{align*}
and if $\distF(f_{\surveyset},\optimalf) \geq \K$, we have:
    \begin{align*}
        \sqrt{\E_{(\covariate,\response) \sim \valD} [(\surveyf(\covariate)-\response)^2]} + \sqrt{\E_{(\covariate,\response) \sim \D} [(\optimalf(\covariate)-\response)^2]} \geq \K  
    \end{align*}
Furthermore, if 
$$\sqrt{\E_{(\covariate,\response) \sim \valD} [(\surveyf(\covariate)-\response)^2]} > \K + \sqrt{\E_{(\covariate,\response) \sim \D} [(\optimalf(\covariate)-\response)^2]}$$
then we have:
\begin{align*}
    \distF(f_{\surveyset},\optimalf) > \K
\end{align*}

\end{lemma}

\begin{lemma}\label{Lemma: Upper Bound for Generalization Error without DP constraint}
    Given a survey sample $\surveyset$ with $|\surveyset| = m$ under \textbf{Assumption 1} and \textbf{Assumption 2} the regression function  $f_{\surveyset}$
    with probability $1 - \frac{\confidence}{2}$ satisfies: 
    \begin{align*}
        \E_{(\covariate,\response) \sim \D}[(\optimalf(\covariate)-\response)^2] = \varnoise \leq \estimatevarS
    \end{align*}
    Where $\estimatevarS = \frac{1}{\sizeS}\sum_{i \in [\sizeS]} (f_{\surveyset}(\covariate_i)-\response_i)^2 + \frac{8\ybound\boundonx\linweight^2\sqrt{2\log{(2d)}}}{\sqrt{\sizeS}} + 3\ybound\sqrt{\frac{\log\frac{4}{\confidence}}{2\sizeS}}$.
\end{lemma}

\begin{proof}
        We know that $\optimalf \in \blinfunc$ is the most optimal linear regression model to fit observations distributed according to $\D$, while $f_{\surveyset} \in \blinfunc$ is the linear regressor learned from a given survey sample $\surveyset$ of size $\sizeS$. 
        

        As per Assumption \ref{Assumption : Assumptions on the linear regression model}, the optimal linear regression model $\optimalf$ is : $\response = 
        \optimalf(\covariate) + \regnoise$, where $\regnoise$ is the zero-mean additive noise term with variance $\Var_{(\covariate,\response) \sim \D} (\regnoise) = \varnoise$. Consequently, we have the following:

        \begin{align}
            \E_{(\covariate,\response) \sim \D} [(\optimalf(\covariate) - \response)^2] = \E_{(\covariate,\response) \sim \D} [\regnoise^2] = \varnoise \label{Eq: Exp of optimal validation function is noise}
        \end{align}

Since the range $\yrange$ of response variable $\response$ is bounded by $[-\ybound,\ybound]$ and  $\optimalf \in \blinfunc$, the Lipschitz constant $\lip$ of the least squares loss can be obtained as : $\lip \leq \sup_{\covariate} 2\ltwonorm{\nabla_{\covariate}\optimalf(\covariate)}|\optimalf(\covariate) - \response| = 2\ybound\ltwonorm{\coeff} \leq 2\ybound\norm{\coeff}_1 \leq 4\ybound\linweight$. Now consider the function $\optimalsurveyf = \argmin_{\sampD} E_{(\covariate,\response) \sim \sampD}[(\surveyf(\covariate)-\response)^2]$. 
Hence, using the value of $\lip$ and  ~\ref{Lemma: Rad-Comp-free generalization bound} and Assumption ~\ref{Assumption : Assumptions on the linear regression model}, we obtain with probability at least $1-\frac{\confidence}{2}$:

        \begin{align}
            &\nonumber\varnoise \\
            = &\nonumber\E_{(\covariate,\response) \sim \sampD}[(\optimalsurveyf(\covariate)-\response)^2]\\
            \leq &\nonumber\E_{(\covariate,\response) \sim \sampD}[(\surveyf(\covariate)-\response)^2] \\         
            \leq &\frac{1}{\sizeS}\sum_{i \in [\sizeS]} (f_{\surveyset}(\covariate_i)-\response_i)^2 + \frac{8\ybound\boundonx\linweight^2\sqrt{2\log{(2d)}}}{\sqrt{\sizeS}}
            + 3\ybound\sqrt{\frac{\log\frac{4}{\confidence}}{2\sizeS}}\label{Eq: Exp of optimal survey function is noise}
        \end{align}


        Combining Equation ~\ref{Eq: Exp of optimal survey function is noise} and ~\ref{Eq: Exp of optimal validation function is noise}, we have with probability at least $1-\frac{\confidence}{2}$:
        \begin{align*}
            \E_{(\covariate,\response) \sim \D} [(\optimalf(\covariate) - \response)^2] \leq &\nonumber\frac{1}{\sizeS}\sum_{i \in [\sizeS]} (f_{\surveyset}(\covariate_i)-\response_i)^2
            + \frac{8\ybound\boundonx\linweight^2\sqrt{2\log{(2d)}}}{\sqrt{\sizeS}} + 3\ybound\sqrt{\frac{\log\frac{4}{\confidence}}{2\sizeS}}
        \end{align*}


    \end{proof}

\subsection{Proof of  Theorem \ref{Theorem: Tester Accepts or Rejects}: Correctness of \SurVerif{}}

\begin{proof}[\textbf{Proof of Theorem \ref{Theorem: Tester Accepts or Rejects}}]

We'll be using the results obtained from Lemma \ref{Lemma: Separation in $l_2$ distance} and Lemma \ref{Lemma: Upper Bound for Generalization Error without DP constraint} to provide the proof for Theorem \ref{Theorem: Tester Accepts or Rejects} .

  First we will show that $\estimatevarD$ is an unbiased estimator of the expected loss $\E_{(\covariate,\response) \sim \D} [(f_{\surveyset}(\covariate) - \response)^2]$ of $f_{\surveyset}$ over the distribution $\D$.

    \begin{align*}
        \E_{(\covariate,\response) \sim \D} [\estimatevarD] = &\E_{(\covariate,\response) \sim \D} [\frac{1}{\sampleD} \sum^{\sampleD}_{i=1} (f_{\surveyset}(\covariate_i) - \response_i)^2] \\
        = &\frac{1}{\sampleD} \sum^{\sampleD}_{i=1}\E_{(\covariate,\response) \sim \D} [(f_{\surveyset}(\covariate_i) - \response_i)^2] \\
        = &\E_{(\covariate,\response) \sim \D} [(f_{\surveyset}(\covariate_i) - \response_i)^2]\\
    \end{align*}

    Note that $\sampleD\estimatevarD = \sum^{\sampleD}_{i=1} (f_{\surveyset}(\covariate_i) - \response_i)^2$ is a sum of $\sampleD$ independent random variables, each of them taking values in the range $[0,4\ybound^2]$; this follows from Lemma~\ref{lem:xyW} and \textbf{Assumption 2}. The application of Hoeffding's inequality gives us the following

    \begin{align*}
        &\Pr[|\estimatevarD - \E_{(\covariate,\response) \sim \valD} [(\surveyf(\covariate)-\response)^2]| \geq 4\tol\boundony] \\
        = &\Pr[|\sampleD\estimatevarD - \sampleD\E_{(\covariate,\response) \sim \D}[\estimatevarD]| \geq 4\sampleD\tol\boundony] \
        \leq 2e^{\frac{-2\sampleD\tol^2}{\ybound^2}}
        \leq \frac{\confidence}{2}
    \end{align*}



Where the last inequality follows from fixing $\sampleD = \frac{\ybound^2\log (\frac{4}{\confidence})}{2\tol^2}$. Thus, if the number of samples $\sampleD$ obtained from $\D$ is at least $\frac{\ybound^2\log (\frac{4}{\confidence})}{2\tol^2}$, we obtain the following bound with probability at least $1 - \frac{\confidence}{2}$, we have:




\begin{align}
    \nonumber&\Pr\left[|\sqrt{\estimatevarD} - \sqrt{\E_{(\covariate,\response) \sim \valD} [(\surveyf(\covariate)-\response)^2]}| \geq \tol\right]\\
    \nonumber\leq &\Pr\Bigg[|\sqrt{\estimatevarD} - \sqrt{\E_{(\covariate,\response) \sim \valD} [(\surveyf(\covariate)-\response)^2]}| \geq\frac{4\boundony\tol}{|\sqrt{\estimatevarD} + \sqrt{\E_{(\covariate,\response) \sim \valD} [(\surveyf(\covariate)-\response)^2]}|}\Bigg]\\
    \nonumber\leq &\Pr\left[|\estimatevarD - (\sqrt{\E_{(\covariate,\response) \sim \valD} [(\surveyf(\covariate)-\response)^2]})^2| \geq 4\boundony\tol\right] \\
    \leq &\frac{\confidence}{2}\label{Eq: Sample Complexity Bound}
\end{align}

Here, we have used the fact $|\sqrt{\estimatevarD} + \sqrt{\E_{(\covariate,\response) \sim \valD} [(\surveyf(\covariate)-\response)^2]}| \leq 4\boundony$.

\paragraph{Proof of First Part of Theorem~\ref{Theorem: Tester Accepts or Rejects}}
    
    Next, we will first show that if $\distF(\optimalf,f_{\surveyset}) \leq \K$
    then \SurVerif{} outputs ACCEPT with probability $1 -\confidence$. 

    Recall that the estimate of deviation of the loss according to the sampling distribution is given by-

    \begin{align*}
        \estimatevarS = \frac{1}{\sizeS}\sum_{i \in [\sizeS]} (f_{\surveyset}(\covariate_i)-\response_i)^2 &+ \frac{8\ybound\boundonx\linweight^2\sqrt{2\log{(2d)}}}{\sqrt{\sizeS}} + 3\ybound\sqrt{\frac{\log\frac{2}{\confidence}}{2\sizeS}},
    \end{align*}
    and the estimate of deviation of the loss according to the validation or test distribution is given by
    \begin{align*}
        \estimatevarD = \frac{1}{\sampleD}\sum^{\sampleD}_{i=1} (f_{\surveyset}(\covariate_i) - \response_i)^{2}.
    \end{align*}

    If $\distF(\optimalf,f_{\surveyset}) \leq \K$, then by applying Lemma ~\ref{Lemma: Separation in $l_2$ distance} followed by Lemma~\ref{Lemma: Upper Bound for Generalization Error without DP constraint} and using the inequality $\sigma^2 \leq \estimatevarS$ we obtain the following with probability at least $1 - \frac{\confidence}{2}$
    
    \begin{align}\label{First equation for completenes}
         \sqrt{\E_{(\covariate,\response) \sim \valD} [(\surveyf(\covariate)-\response)^2]} \leq \K + \sqrt{\estimatevarS}
    \end{align}
    
    Given $\sqrt{\E_{(\covariate,\response) \sim \valD} [(\surveyf(\covariate)-\response)^2]} \leq \K + \sqrt{\estimatevarS}$, observe that from 
    Eq. ~\ref{Eq: Sample Complexity Bound}, we obtain the following with probability at least $1 - \frac{\confidence}{2}$,
    \begin{align}\label{Second equation for completeness}
        \sqrt{\estimatevarD} \leq \sqrt{\estimatevarS} + \K + \tol
    \end{align}

    Combining Equations \eqref{First equation for completenes} and \eqref{Second equation for completeness}, if $\distF(f_{\surveyset}, \response) \leq \K + \sqrt{\estimatevarS}$ then 
    we obtain the following guarantee with probability at least $(1 - \frac{\confidence}{2})^2 \geq (1 - \confidence)$:
    \begin{align*}
        \sqrt{\estimatevarD} \leq \sqrt{\estimatevarS} + \K + \tol  
    \end{align*}

    Thus, if we have $\distF(\surveyf,\optimalf)\leq \K$, our tester \SurVerif{} ACCEPTS with probability at least $1 - \confidence$.

\paragraph{Proof of Second Part of Theorem~\ref{Theorem: Tester Accepts or Rejects}}

    If the Algorithm \SurVerif{} rejects, we have:
    \begin{align*}
        \sqrt{\estimatevarD} > \sqrt{\estimatevarS} + \K + \tol  
    \end{align*}
    Then, by Lemma~\ref{Lemma: Upper Bound for Generalization Error without DP constraint}, we have with probability $1 - \frac{\confidence}{2}$:
    \begin{align*}
        \sqrt{\estimatevarD} > \sqrt{\E_{(\covariate,\response) \sim \D} [(\optimalf(\covariate)-\response)^2]}  + \K + \tol
    \end{align*}
    Consequently, from Equation~\ref{Eq: Sample Complexity Bound}, we have with probability $1- \confidence$,
    \begin{align*}
        \small{\sqrt{\E_{(\covariate,\response) \sim \D} [(\surveyf(\covariate)-\response)^2]} > \sqrt{\E_{(\covariate,\response) \sim \D} [(\optimalf(\covariate)-\response)^2]} + \K }
    \end{align*}
    Consequently, by Lemma~\ref{Lemma: Separation in $l_2$ distance}, we have:
    \begin{align*}
        \distF(f_{\surveyset},\optimalf) > \K
    \end{align*}

\paragraph{Proof of Third Part of Theorem~\ref{Theorem: Tester Accepts or Rejects}}
Given $|\surveyset| = \sizeS \rightarrow \infty$, we have:
\begin{align*}
   \estimatevarS = \frac{1}{\sizeS} \sum_{i \in [\sizeS]} (\surveyf(\covariate_i) - \response_i)^2 \rightarrow \E_{(\covariate,\response) \sim \D} [(\surveyf(\covariate)-\response)^2] 
\end{align*}
This ensures $\surveyf \rightarrow \optimalsurveyf$, and hence $\estimatevarS \rightarrow \varnoise$. Then, we have:
\begin{align*}
    \E_{(\covariate,\response) \sim \valD} [(\surveyf(\covariate)-\response)^2] = \estimatevarS = \varnoise
\end{align*}
If $\distF(\surveyf,\optimalf) \geq \K + 2\sdnoise + 2\tol$, by Lemma~\ref{Lemma: Separation in $l_2$ distance} we have :
\begin{align*}
    \sqrt{\E_{(\covariate,\response) \sim \valD} [(\surveyf(\covariate)-\response)^2]} + \sqrt{\E_{(\covariate,\response) \sim \D} [(\optimalf(\covariate)-\response)^2]} \geq \K + 2\sdnoise + 2\tol
\end{align*}
Hence, using Equation~\ref{Eq: Sample Complexity Bound}, we have with probability $1 - \confidence$,
\begin{align*}
    \sqrt{\estimatevarD} + \tol + \sqrt{\estimatevarS} &\geq \K + 2\sdnoise + 2\tol\\
    \sqrt{\estimatevarD} &\geq \K + \sdnoise + \tol
\end{align*}
This concludes our proof.
\end{proof}
\newpage
\section{Generalization of \SurVerif{} to \dppair-LDP and $\dpepsilon$-LDP data}

\begin{lemma}\label{Lemma: Upper Bound for Generalization Error with DP}
    Given a data set $\surveyset$ generated from a distribution $\sampD$ s.t. any $(\covariate, \response) \sim \sampD$ generated satisfies a linear model $\response = \inn{\coeff^*}{\covariate} + \regnoise$. Also, given an empirical $\coeffhat$ such that $\norm{\coeffhat - \coeff^*} \leq \coeffdist$ with probability at least $1 - \confidence_1$ and $\E_{(\covariate,\response) \sim \sampD} \left[(\response - \inn{\coeffhat}{\covariate})^2\right] \leq \estimatevarS$ with probability at least $1 - \confidence_2$. Then, we have with probability at least $1 - \confidence_1 -  \confidence_2$,
    \begin{align*}
        \standevnoise \leq \sqrt{\estimatevarS} + \boundonx\coeffdist
    \end{align*}
\end{lemma}
    \begin{proof}

    Standard Deviation of noise is given by
        \begin{align*}
        \standevnoise &= \sqrt{\E_{(\covariate,\response) \sim \sampD} \left[(\response-\inn{\coeff^*}{\covariate})^2\right]}\\
        &\leq \sqrt{\E_{(\covariate,\response) \sim \D} \left[(\response - \inn{\coeffhat}{\covariate})^2\right]} + \sqrt{\E_{(\covariate,\response) \sim \sampD}(\inn{\coeffhat- \coeff^*}{\covariate})^2}\\
        &\leq \sqrt{\E_{(\covariate,\response) \sim \D} \left[(\response - \inn{\coeffhat}{\covariate})^2\right]} + \boundonx\sqrt{\E_{(\covariate,\response) \sim \sampD}\norm{\coeffhat- \coeff^*}^2_2}
    \end{align*}
    With a probability of at least  $1 - \confidence_2$, we have
    \begin{align*}
        \standevnoise \leq \sqrt{\estimatevarS} + \boundonx\sqrt{\E_{(\covariate,\response) \sim \sampD}\norm{\coeffhat- \coeff^*}^2_2}
    \end{align*}
    With probability at most $\confidence_2$, we have 
    \[
    \standevnoise > \sqrt{\estimatevarS} + \boundonx\sqrt{\E_{(\covariate,\response) \sim \sampD}\norm{\coeffhat- \coeff^*}^2_2}
    \]
    and with probability almost $\confidence_1$, we have
    \[
    \norm{\coeffhat - \coeff^*}_2 > \coeffdist
    \]
    Using union bound on the above two events, We have our result.
    \end{proof}

Lemma \ref{Lemma: Upper Bound for Generalization Error with DP} establishes a general algorithmic framework to extend \SurVerif{} to \dppair-LDP and $\dpepsilon$-LDP data, where we get an extra error term  $\coeffdist$ due to the local privacy constraints. In the following algorithms, we obtain an estimate of $\coeffdist$ as $\hat{J}$ and use it to obtain a generalization error that holds with high probability for LDP data. The algorithms retains the original algorithmic template and the sample complexity of \SurVerif{}. The only changes are highlighted in {\color{blue}blue}.  We also note that we assume the additional samples obtained from $\D$ to not be subject to LDP constraints.

\subsection{\PriVerif: Extension of \SurVerif{} to $(\alpha, \beta)$-LDP}
\begin{algorithm}[h!]
\caption{\PriVerif($\surveyset \subset \mathbb{R}^{(d+1)},\D,\K,\confidence,\tol, \ybound, \linweight, \dpepsilon,  \dpdelta, \boundonx$)}\label{alg:PriVerif_alpha-beta-private-private}
\begin{algorithmic}[1]
\State Initialize $\sampleD \gets \ceil*{\frac{\ybound^2\log (\frac{4}{\confidence})}{2\tol^2}}$, $\sizeS \gets |\surveyset|$, $\surveyset_{\D} \gets \emptyset$
\State $(\dpsurveyset,\noisevariance) \gets \AlgoPriv\left(\surveyset,\dpepsilon,\dpdelta,\boundonx\right)$
\State $\coeffhat \gets \AlgoReg(\dpsurveyset,\noisevariance, \lassobound)$
\State $\hat{L} \gets \frac{1}{\sizeS}\sum_{(\covariate,\response) \in \surveyset} (f_{\surveyset}(\covariate) - \response)^2$, where $f_{\surveyset}(\covariate)= \inn{\coeffhat}{\covariate}$
\State {\color{blue}$\hat{J} \gets \frac{2\constant_2 \boundonx^3}{\lambda_{\min}(\xvariance)}     \frac{\sqrt{\log\left(\frac{1}{\dpdelta}\right)}}{\dpepsilon}\left(\frac{\log\left(\frac{1}{\dpdelta}\right)}{\dpepsilon}+1\right)
     \linweight\sqrt{\frac{d\log{d}}{\sizeS}}$} \label{PriVerif Line 5}
\State $\estimatevarS \gets \hat{L} + \frac{8\ybound\boundonx\linweight^2\sqrt{2\log{(2d)}}}{\sqrt{\sizeS}} +  3\ybound\sqrt{\frac{\log\frac{4}{\confidence}}{2\sizeS}}
{\color{blue}+ \hat{J}}
$
\State $\surveyset_{\D} \gets \sampleD \text{ iid samples from } \D$. 
\State $\estimatevarD \gets \frac{1}{\sampleD}\sum_{(\covariate,\response) \in \surveyset_{\D}} (f_\surveyset(\covariate) - \response)^2$ 
\If{$\sqrt{\estimatevarD} > \sqrt{\estimatevarS} + \K + \tol  $}~~Output REJECT. 
\Else ~~Output ACCEPT.
\EndIf
\end{algorithmic}
\end{algorithm}

\begin{lemma}\label{Lemma: Upper Bound for Generalization Error with (alpha,beta)-DP constraints}
   Let us consider a survey data $\surveyset$ with  $\sizeS \geq  \max\left(\frac{
    \constant}{\lambda^2_{min}(\xvariance)}\left(\boundonx^2+\frac{\boundonx^2 \log{\left(\frac{1}{\dpdelta}\right)}}{\dpepsilon^2}\right)^2d\log{d}, 1\right)$ samples generated from a linear model $\response = \inn{\coeff}{\covariate} + \regnoise$ satisfying $\norm{\coeff^*}_1 \leq \linweight$, and $|\covariatei_i|\leq \boundonx, \forall i \in [d]$, where $\regnoise$ comes from a subgaussian distribution with parameter $\sigma_\regnoise$. Now, if we apply \emph{\AlgoPriv} satisfying \dppair-local DP on $\surveyset$ where $\dpepsilon \leq \frac{\boundonx \sqrt{\log\left(\frac{1}{\dpdelta}\right)}}{\standevnoise}$ , then run \emph{\AlgoReg} to obtain $\coeffhat$. Then, for some constants $c_1$ and $c_2$, with probability at least  $1-\confidence-d^{-\constant_1}$ we have:

    \begin{align*}
    \standevnoise~~\leq &\sqrt{\frac{1}{\sizeS}\sum_{i \in [\sizeS]} (\response_i - \inn{\coeffhat}{\covariate_i})^2 + \frac{8\ybound\boundonx\linweight^2\sqrt{2\log{(2d)}}}{\sqrt{\sizeS}} + 3\ybound\sqrt{\frac{\log\frac{2}{\confidence}}{2\sizeS}}} \\
    &+ 
     \frac{2\constant_2 \boundonx^3}{\lambda_{\min}(\xvariance)}
     \frac{\sqrt{\log\left(\frac{1}{\dpdelta}\right)}}{\dpepsilon}\left(\frac{\log\left(\frac{1}{\dpdelta}\right)}{\dpepsilon}+1\right)
     \linweight\sqrt{\frac{d\log{d}}{\sizeS}}
    \end{align*}
\end{lemma}
\begin{proof}
    
From Theorem \ref{Theorem: Learning from Laplacian LDP data}, using a survey data $\surveyset$ with $\sizeS$ samples, the linear model $\coeffhat$ learnt from \AlgoReg{} satisfies with probability at least $1-d^{-\constant_1}$ and using $\dpepsilon \leq \frac{\boundonx \sqrt{\log\left(\frac{1}{\dpdelta}\right)}}{\standevnoise}$ we get,

\begin{align*}
    \norm{\coeff^*-\coeffhat}_2 \leq \frac{2\constant_2\linweight\boundonx^2}{\lambda_{\min}(\xvariance)}
    \frac{\sqrt{\log\left(\frac{1}{\dpdelta}\right)}}{\dpepsilon}\left(\frac{\log\left(\frac{1}{\dpdelta}\right)}{\dpepsilon}+1\right)
     \linweight\sqrt{\frac{d\log{d}}{\sizeS}}
\end{align*}

From Lemma \ref{Lemma: Rad-Comp-free generalization bound} we obtain with probability at least $1 - \confidence$,
\begin{align*}
    \E_{(\covariate,\response) \sim \sampD} \left[(\response - \inn{\coeffhat}{\covariate})^2\right] \leq \frac{1}{\sizeS}\sum_{i \in [\sizeS]} (f_{\surveyset}(\covariate_i)-\response_i)^2 + \frac{8\ybound\boundonx\linweight^2\sqrt{2\log{(2d)}}}{\sqrt{\sizeS}} + 3\ybound\sqrt{\frac{\log\frac{2}{\confidence}}{2\sizeS}}
\end{align*}
Combining the above results using Lemma \ref{Lemma: Upper Bound for Generalization Error with DP}, we obtain our result with probability at least  $1-\confidence-d^{-\constant_1}$.

\end{proof}
Using Lemma \ref{Lemma: Upper Bound for Generalization Error with (alpha,beta)-DP constraints} we have a testing framework \PriVerif{} for \dppair-LDP. In \PriVerif{} the additional error term is $\hat{J} = \frac{2\constant_2 \boundonx^3}{\lambda_{\min}(\xvariance)}     \frac{\sqrt{\log\left(\frac{1}{\dpdelta}\right)}}{\dpepsilon}\left(\frac{\log\left(\frac{1}{\dpdelta}\right)}{\dpepsilon}+1\right)
     \linweight\sqrt{\frac{d\log{d}}{\sizeS}} $ in Line \ref{PriVerif Line 5}.

\subsection{\PriVerifL: Extension of \SurVerif{} to $\alpha$-LDP}

\begin{algorithm}[h!]
\caption{\PriVerifL($\surveyset \subset \mathbb{R}^{(d+1)},\D,\K,\confidence,\tol, \ybound, \linweight, \dpepsilon,  \boundonx$)}\label{alg:PriVerif_alpha-beta-private-private}
\begin{algorithmic}[1]
\State Initialize $\sampleD \gets \ceil*{\frac{\ybound^2\log (\frac{4}{\confidence})}{2\tol^2}}$, $\sizeS \gets |\surveyset|$, $\surveyset_{\D} \gets \emptyset$
\State $(\dpsurveyset,\noisevariance) \gets \AlgoPriv\left(\surveyset,\dpepsilon,0,\boundonx\right)$
\State $\coeffhat \gets \AlgoReg(\dpsurveyset,\noisevariance, \lassobound)$
\State $\hat{L} \gets \frac{1}{\sizeS}\sum_{(\covariate,\response) \in \surveyset} (f_{\surveyset}(\covariate) - \response)^2$, where $f_{\surveyset}(\covariate)= \inn{\coeffhat}{\covariate}$
\State {\color{blue}$\hat{J} \gets \frac{\constant_2\boundonx}{\lambda_{\min}(\xvariance)}\max\left(\frac{\boundonx}{\dpepsilon},\boundonx^2, \constant_\regnoise\right) \linweight\sqrt{\frac{d\log{d}}{\sizeS}}$} \label{PreVerifL line 5}
\State $\estimatevarS \gets \hat{L} + \frac{8\ybound\boundonx\linweight^2\sqrt{2\log{(2d)}}}{\sqrt{\sizeS}} +  3\ybound\sqrt{\frac{\log\frac{4}{\confidence}}{2\sizeS}}
{\color{blue}+ \hat{J}}
$
\State $\surveyset_{\D} \gets \sampleD \text{ iid samples from } \D$. 
\State $\estimatevarD \gets \frac{1}{\sampleD}\sum_{(\covariate,\response) \in \surveyset_{\D}} (f_\surveyset(\covariate) - \response)^2$ 
\If{$\sqrt{\estimatevarD} > \sqrt{\estimatevarS} + \K + \tol  $}~~Output REJECT. 
\Else ~~Output ACCEPT.
\EndIf
\end{algorithmic}
\end{algorithm}

\begin{lemma}\label{Lemma: Upper Bound for Generalization Error with Local DP constraints}
    Consider a survey data  containing $\surveyset$ with $\sizeS \geq \max\bigg( \max\left(\frac{\max\left(\frac{\boundonx}{\dpepsilon},\boundonx^2, \constant_\regnoise\right)}{\lambda_{\min}(\xvariance)},1\right)d\log{d}, 
 \max\left(\frac{\boundonx}{\dpepsilon},\boundonx^2, \constant_\regnoise\right) \log^3(d)\bigg)$ data points $(\covariate,\response)$ samples generated from a linear model $\response = \inn{\coeff^*}{\covariate} + \regnoise$ satisfying  $\norm{\coeff^*}_1 \leq \linweight$, $|\covariatei_i|\leq \boundonx, \forall i \in [d]$, where $\regnoise$ comes from a sub-exponential distribution such that $\Pr[\regnoise \geq t] \leq \exp\left(-\frac{t}{\constant_\regnoise}\right)$. Now, if we apply \emph{\AlgoPriv} satisfying $\dpepsilon$-local DP on $\surveyset$, then run \emph{\AlgoReg} to obtain $\coeffhat$. Then, for some constants $c_1$ and $c_2$, with probability at least $1-\confidence-d^{-\constant_1}$ we have, 
    
  \begin{align*}
    \standevnoise \leq &\sqrt{\frac{1}{\sizeS}\sum_{i \in [\sizeS]} (\response_i - \inn{\coeffhat}{\covariate_i})^2 + \frac{8\ybound\boundonx\linweight^2\sqrt{2\log{(2d)}}}{\sqrt{\sizeS}} + 3\ybound\sqrt{\frac{\log\frac{2}{\confidence}}{2\sizeS}}} \\
    &+ 
    \frac{\constant_2\boundonx}{\lambda_{\min}(\xvariance)}\max\left(\frac{\boundonx}{\dpepsilon},\boundonx^2, \constant_\regnoise\right) \linweight\sqrt{\frac{d\log{d}}{\sizeS}}
\end{align*}
\end{lemma}

\begin{proof}
From Theorem \ref{Theorem: Learning from Gaussian LDP data}, using a survey data $\surveyset$ with $\sizeS$ samples, the linear model $\coeffhat$ learned from \AlgoReg{} staisfies with probability at least $1-d^{-\constant_1}$:

    \begin{align*}
    \norm{\coeff^*-\coeffhat}_2 \leq
        \frac{\constant_2}{\lambda_{\min}(\xvariance)}\max\left(\frac{\boundonx}{\dpepsilon},\boundonx^2, \constant_\regnoise\right)\linweight\sqrt{\frac{d\log{d}}{\sizeS}}\,\\
\end{align*}

From Lemma \ref{Lemma: Rad-Comp-free generalization bound} we obtain with probability at least $1 - \confidence$,
\begin{align*}
    \E_{(\covariate,\response) \sim \sampD} \left[(\response - \inn{\coeffhat}{\covariate})^2\right] \leq \frac{1}{\sizeS}\sum_{i \in [\sizeS]} (f_{\surveyset}(\covariate_i)-\response_i)^2 + \frac{8\ybound\boundonx\linweight^2\sqrt{2\log{(2d)}}}{\sqrt{\sizeS}} + 3\ybound\sqrt{\frac{\log\frac{2}{\confidence}}{2\sizeS}}
\end{align*}
    Combining the above results using Lemma \ref{Lemma: Upper Bound for Generalization Error with DP}, we obtain our result with probability at least  $1-\confidence-d^{-\constant_1}$.
\end{proof}

Using Lemma \ref{Lemma: Upper Bound for Generalization Error with Local DP constraints} we have a testing framework \PriVerifL{} for $\alpha$-LDP. In \PriVerifL{} the additional error term is $\hat{J} = \frac{\constant_2\boundonx}{\lambda_{\min}(\xvariance)}\max\left(\frac{\boundonx}{\dpepsilon},\boundonx^2, \constant_\regnoise\right) \linweight\sqrt{\frac{d\log{d}}{\sizeS}}$ in .

\clearpage
\section{Proofs of Section 4: Surveys with LDP}
\paragraph{\dppair-Local Differential Privacy.}
\begin{lemma}[Learning from data with additive subgaussian noise~\cite{LDP_Regression_AOS}]\label{lemma: Additive Subgaussian noise regression}
    Given a linear regression problem of the form $\response = \inn{\coeff}{\covariate} + \regnoise, \covariate \in \R^d$ with optimal solution $\coeff^*$ where we observe $\noisycovariate = \covariate + \noisevariable$ and $\covariate,\noisevariable$ are $\sigma_\covariate^2,\sigma_\noisevariable^2$ (resp.) subgaussian random vectors, the algorithm \ref{Algorithm: LDP Survey Regression} learns $\coeffhat$ using $\sizeS \geq \max{\left(\frac{(\sigma_\covariate^2+\sigma_\noisevariable^2)^2}{\lambda^2_{\min}(\xvariance)},1\right)}d\log{d}$ samples such that for some constants $\constant_1,\constant_2$, with probability at least $1 - d^{-\constant_1}$, we have:
    \[\norm{\coeffhat-\coeff^*}_2 \leq \constant_2 \frac{\sigma_\noisycovariate(\sigma_\noisevariable+\sigma_\regnoise)}{\lambda_{\min}(\xvariance)}\norm{\coeff^*}_2\sqrt{\frac{d\log{d}}{\sizeS}}\]
\end{lemma}

We use the Lemma \ref{lemma: Gaussian LDP Survey} and \ref{lemma: Additive Subgaussian noise regression} to prove the Theorem \ref{Theorem: Learning from Gaussian LDP data}:

\begin{proof}[Proof of Theorem \ref{Theorem: Learning from Gaussian LDP data}]
    Under the conditions stated in the theorem, \ref{lemma: Gaussian LDP Survey} ensures that $\AlgoPriv$ outputs $\surveyset'$ satisfying \dppair-privacy and the variance of noise $\Sigma_\noisevariable = \frac{\constant_1\boundonx\sqrt{\log\frac{1}{\dpdelta}}}{\dpepsilon}\identity_d$. This ensures that the noise $\noisevariable$ is a $\frac{\constant_1\boundonx\sqrt{\log\frac{1}{\dpdelta}}}{\dpepsilon}$-subgaussian random vector. The random vector $\covariate$ is bounded as $|\covariate_i|\leq \boundonx$, and hence $2\boundonx$-subgaussian. This gives us the upper bound on $\norm{\coeff^*-\coeffhat}_2$ as:
    
    \[\constant_2\frac{\boundonx\sqrt{\frac{\log\left(\frac{1}{\dpdelta}\right)}{\dpepsilon}+1}\left(\frac{\boundonx\sqrt{\log\left(\frac{1}{\dpdelta}\right)}}{\dpepsilon}+\sigma_\regnoise\right)}{\lambda_{\min}(\xvariance)}\norm{\coeff^*}_2\sqrt{\frac{d\log{d}}{\sizeS}}\]
    
    Putting in the upper bound on $\norm{\coeff^*}_2$ as $\linweight$ gives the bound.
\end{proof}


\paragraph{$\dpepsilon$-Local Differential Privacy.}
We state the following result from ~\cite{LDP_Regression_AOS}:

\begin{theorem}\label{Theorem: Larger Deviation Noisy Regression}
    Given a linear regression problem of the form $\response = \inn{\coeff}{\covariate} + \regnoise, \covariate \in \R^d$ with optimal solution $\coeff^*$ where we observe $\noisycovariate = \covariate + \noisevariable$, Algorithm \ref{Algorithm: LDP Survey Regression} satisfies the following bounds if there exists  $\Phi(\noisevariable)$ such that $\norm{\predcovardummy - \basecovardummy\coeff^*}_\infty \leq \Phi(\noisevariable)\sqrt{\frac{\log(d)}{\sizeS}}$, and the matrix $\basecovardummy$ satisfies the lower restricted eigenvalue (lower-RE) condition for all $\coeff \in \R^d$, $\coeff^T\basecovardummy\coeff \geq \lowerrealpha \norm{\coeff}_2^2-\tau(\sizeS,d)\norm{\coeff}_1^2$ for some $\lowerrealpha > 0$ and $\tau(\sizeS,d) > 0$ with $\tau(\sizeS,d) \leq \frac{\lowerrealpha}{2d}$:
    \begin{align}
        \norm{\coeffhat-\coeff^*}_2 \leq \frac{\constant}{\lowerrealpha}\Phi(\noisevariable)\sqrt{\frac{d\log{d}}{\sizeS}}
    \end{align}
    Where $\constant$ is a constant.
\end{theorem}

Observe that the Theorem \ref{Theorem: Larger Deviation Noisy Regression} is a deterministic result. We would like to show that the conditions stated in the theorem hold with high probability in the case for \AlgoPriv{} with $\dpepsilon$-LDP. We formalize this idea in the following Theorem ~\ref{Theorem: Sub-Expoential Noise Noisy Regression}. For that purpose, we use the following results from \cite{LDP_Regression_AOS}.


\begin{lemma}\label{lemma: RE conditions reduction}
    Suppose $s \geq 1$ and $\basecovardummy$ is an estimator of $\Sigma_x$ satisfying the deviation conditions:
    \begin{align*}|\coeff^T(\basecovardummy-\xvariance)\coeff| &\leq \frac{\lambda_{\min}(\xvariance)}{54},\qquad\forall \coeff \in \bK(2s)
    \end{align*}
    Then we have the lower-RE condition:
    \[\coeff^T\basecovardummy\coeff \geq \frac{\lambda_{\min}(\xvariance)}{2}\norm{\coeff}_2^2-\frac{\lambda_{\min}(\xvariance)}{2s}\norm{\coeff}_1^2\]
    Where $\lambda_{\min}(\mathbf{M})$ denotes the least eigenvalues of the matrix $M$. And $\bK(2s):= \bfB_0(s)\cap \bfB_2(1)$ where $\bfB_p(r)$ denote balls of radius $r$ in $p$-dimensional box.
\end{lemma}

\begin{lemma}\label{lemma: ball bound laplace}
    If $\covariatematrix \in \R^{\sizeS\times d}$ is a random matrix whose each entry are i.i.d. such that $\Pr(x_i \geq t) \leq \exp\left(-\frac{t}{\constant_\covariate}\right)$, then there is a universal constant $\constant$ such that:
    \begin{align*}
    &\Pr\left[\sup_{\coeff \in \bK(2s)}\left|\frac{\norm{\covariatematrix\coeff}_2^2}{\sizeS}-\E\left[\frac{\norm{\covariatematrix\coeff}_2^2}{\sizeS}\right]\right|\geq t\right] \leq \constant\exp\left(-\frac{\sizeS t^2}{\constant_\covariate^2}+2s\log d\right)\,.
    \end{align*}
\end{lemma}

\begin{proof}
    The proof follows the proof of Lemma 15 of ~\cite{LDP_Regression_AOS} with the use of the inequality stated in Lemma ~\ref{Lemma: Laplace Squared Tail}.
\end{proof}

Now, we use these results to show that the assumptions stated in Theorem \ref{Theorem: Larger Deviation Noisy Regression} hold with high probability under the conditions stated in Theorem \ref{Theorem: Sub-Expoential Noise Noisy Regression}:


\begin{lemma}\label{lemma: Restricted Eigenvalue}
    Under the conditions of Theorem \ref{Theorem: Sub-Expoential Noise Noisy Regression}, we have the lower-RE condition satisfied with $\lowerrealpha = \frac{\lambda_{\min}(\xvariance)}{2}$ and $\tau(\sizeS) = \constant_1 \lambda_{\min}(\xvariance)\max\left(\frac{\constant_{\max}^2}{\lambda^2_{\min}(\xvariance)} , 1 \right) \frac{\log d}{\sizeS}$ with probability at least $1-\constant\exp\left(-\sizeS\min\left(\frac{\lambda^2_{\min}(\xvariance)}{\constant_{\max}^2},1\right)\right)$

    \begin{proof}
        By Lemma \ref{lemma: RE conditions reduction}, and the fact that $\basecovardummy - \xvariance = \frac{\noisycovariatematrix^T\noisycovariatematrix}{\sizeS}-\zvariance$, we can fix $s = \frac{1}{\constant_1}\frac{\sizeS}{\log d}\min\left(\frac{\lambda^2_{\min}(\xvariance)}{\constant_{\max}^2},1\right)$ to obtain the stated bound if $\sup_{\coeff\in\bK(2s)} \left|\coeff^T\left(\frac{\noisycovariatematrix^T\noisycovariatematrix}{\sizeS}-\zvariance\right)\coeff\right| \leq \frac{\lambda_{\min}(\xvariance)}{54}$. Hence, it suffices to show that this bound holds with high probability. Note that $\noisycovariatematrix$ satisfies the condition of Lemma \ref{lemma: ball bound laplace} with $\constant_{\max} = \max(\constant_\covariate , \constant_\noisevariable)$.

        \begin{align*}
            \Pr\left[\sup_{\coeff\in\bK(2s)} \left| 
            \coeff^T  \left(\frac{\noisycovariatematrix^T\noisycovariatematrix}{\sizeS}-\zvariance\right)\coeff\right| \geq t\right] 
            \leq &\constant_2 \exp\left(-\frac{\sizeS t^2}{\constant_{\max}^2}+2s\log d\right)\\
            \leq &\constant_2\exp\left(-\sizeS\min\left(\frac{\lambda^2_{\min}(\xvariance)}{\constant_{\max}^2},1\right)\right)
        \end{align*}
        Where the first inequality follows from Lemma \ref{lemma: ball bound laplace} and the second inequality follows from fixing $t = \frac{\lambda_{\min}(\xvariance)}{54}$ and a sufficiently large $\constant_1$.
    \end{proof}
\end{lemma}

\begin{lemma}\label{Lemma: Bounded Deviation}
    Under the conditions of Theorem \ref{Theorem: Sub-Expoential Noise Noisy Regression}, the deviation bound condition on $\norm{\predcovardummy - \basecovardummy\coeff^*}_\infty$ is satisfied by:
    \[
    \Phi(\noisycovariatematrix) \leq \constant_1\constant_{max}\norm{\coeff^*}_2
    \]
    with probability at least $1 - \frac{1}{d^{\constant_2}}$ where $\constant_1,\constant_2$ are constants, and $\constant_{max} = \max(\constant_\covariate,\constant_\noisevariable,\constant_\regnoise)$.

    \begin{proof}
        We use the model assumption $\response = \inn{\coeff}{\covariate} + \regnoise$ to upper bound $\norm{\predcovardummy - \basecovardummy\coeff^*}_\infty$:
        \begin{align*}
            \norm{\predcovardummy - \basecovardummy\coeff^*}_\infty &= \norm{\frac{\noisycovariatematrix^T \responsevector}{\sizeS}-\left(\frac{\noisycovariatematrix^T\noisycovariatematrix}{\sizeS}-\wvariance
        \right)\coeff^*}\\
            &= \norm{\frac{\noisycovariatematrix^T(\covariatematrix\coeff^*+\regnoisevector)}{\sizeS}-\left(\frac{\noisycovariatematrix^T\noisycovariatematrix}{\sizeS}-\noisevariance\right)\coeff^*}_\infty\\
            &= \norm{\frac{\noisycovariatematrix^T\regnoisevector}{\sizeS}-\left(\frac{\noisycovariatematrix^T(\noisycovariatematrix-\covariatematrix)}{\sizeS}-\noisevariance\right)\coeff^*}_\infty\\
            &= \norm{\frac{\noisycovariatematrix^T\regnoisevector}{\sizeS}+\left(\noisevariance-\frac{\noisycovariatematrix^T\noisematrix}{\sizeS}\right)\coeff^*}_\infty\\
            &\leq \norm{\frac{\noisycovariatematrix^T\regnoisevector}{\sizeS}}_\infty + \norm{\left(\noisevariance-\frac{\noisycovariatematrix^T\noisematrix}{\sizeS}\right)\coeff^*}_\infty
        \end{align*}
        Consequently Lemma \ref{lemma: laplace random matrix deviation bound} completes the proof.
    \end{proof}
\end{lemma}

    \begin{proof}[Proof of Theorem \ref{Theorem: Sub-Expoential Noise Noisy Regression}]
        The Lemma \ref{lemma: Restricted Eigenvalue} and \ref{Lemma: Bounded Deviation} ensures that the deviation condition and restricted eigenvalues are satisfied with $\lowerrealpha = \frac{\lambda_{\min}(\xvariance)}{2}$, and $\tau(\sizeS) = \constant_1 \lambda_{\min}(\xvariance)\max\left(\frac{\constant_{\max}^2}{\lambda^2_{\min}(\xvariance)} , 1 \right) \frac{\log d}{\sizeS}$ under the conditions given. Additionally, fixing $\sizeS \geq \max\left\{ \max\{\frac{\constant_{\max}}{\lambda_{\min}(\xvariance)},1\}d\log{d}, \constant_{max} \log^3(d)\right\}$ ensures the success probabilities are at least $1 - \frac{1}{d^{\constant_2}}$, as well as $\tau(\sizeS) \leq \frac{\lowerrealpha}{2d}$.
    \end{proof}

    \begin{proof}[Proof of Theorem \ref{Theorem: Learning from Laplacian LDP data}]
        Under the conditions stated in the theorem, \ref{lemma: Gaussian LDP Survey} ensures that $\AlgoPriv$ outputs $\surveyset'$ satisfying $\dpepsilon$-privacy and the variance of noise $\noisevariance = \frac{8\boundonx^2}{\dpepsilon^2}$. Additionally the noise variables $\noisevariable$ satisfies the condition of Theorem \ref{Theorem: Sub-Expoential Noise Noisy Regression} with $\constant_\noisevariable = \frac{2\boundonx}{\dpepsilon}$. Also, the bounded domain assumption on $\covariate$ ensures it satisfies the conditions with $\constant_\covariate = \boundonx^2$. Putting these values in Theorem \ref{Theorem: Sub-Expoential Noise Noisy Regression}, we get that with probability $1- \frac{1}{d^{\constant_1}}$ , $\norm{\coeff^*-\coeffhat}_2$ is upper bounded by:
        \[
        \frac{\constant_2}{\lambda_{\min}(\xvariance)}\max\left(\frac{\boundonx}{\dpepsilon},\boundonx^2, \constant_\regnoise\right)\norm{\coeff^*}_2\sqrt{\frac{d\log{d}}{\sizeS}}
        \]
        When we have 
        \begin{align*}
            \sizeS\geq \max\bigg( \max\left(\frac{\max\left(\frac{\boundonx}{\dpepsilon},\boundonx^2, \constant_\regnoise\right)}{\lambda_{\min}(\xvariance)},1\right)d\log{d},  
 \max\left(\frac{\boundonx}{\dpepsilon},\boundonx^2, \constant_\regnoise\right) \log^3(d)\bigg)
        \end{align*}
    \end{proof}


\clearpage
\section{Sub-Weibull Tail Bounds}\label{sec:subweibull}
In this section, we use new bounds on tails of Sub-Weibull distributions to introduce deviation bounds for the sum of squares of Laplace random variables. Note that while we prove the bounds for the Laplace distribution, a special case of Sub-Exponential distributions, the techniques can be extended to general Sub-Exponential distributions by fixing the tails appropriately. First, we state a deviation bound on the Sub-Weibull distributions from \cite{Sharp_SubWeibull_Bakhshizadeh}:

\begin{lemma}[\textbf{Right Tail Bound on Sub-weibull Distributions}]\label{Lemma: subweibull tail bound}
    For an i.i.d. sequence of centered random variables $X_i$ whose right tails are captured by $\constant_\alpha\sqrt[\alpha]{t}$, i.e. $\Pr[X_i \geq t] \leq \exp{(\constant_\alpha\sqrt[\alpha]{t})}$ for some $\alpha > 1$, and $\E[X_i^2|\indicator(X_i \leq 0)]  = \sigma^2_-$. Define $S_n = \sum_{i \in [n]} X_i$. Then, for any $n$, any $0 < \beta < 1$, we have
    \begin{align*}
        \Pr[S_n > nt] \leq &\exp\left(-\frac{nt^2}{\sigma^2_- + \constant_1(\beta,\alpha) + (nt)^{\frac{1}{\alpha}-1}\constant_2(\beta,\alpha)}\right) + \exp\left(-\beta \constant_\alpha \sqrt[\alpha]{nt}\right) +n\exp(-\constant_\alpha \sqrt[\alpha]{nt})\,,
    \end{align*}
    where $\constant_1(\beta,\alpha) = \frac{\Gamma(2\alpha+1)}{((1-\beta)\constant_\alpha)^{2\alpha}}$ and $\constant_2(\beta,\alpha) = \frac{\beta \constant_\alpha \Gamma(3\alpha +1)}{3((1-\beta)\constant_\alpha)^{3\alpha}}$.
\end{lemma}

Now, we can fix the tail capturing function for the square of Laplace random variables to obtain the right tail deviation bounds for the square of Laplace random variables.

\begin{lemma}\label{lemma: Subexponential Squared Right Tail}
    Given an i.i.d. sequence $X_i $ such that $\Pr(|X_i| \geq t) \leq \exp\left(-\frac{t}{\constant_x}\right)$ with $\constant_x \geq 1$, define a corresponding i.i.d sequence $Y_i = X_i^2 - \E[X^2]$. Define $S_n = \sum_{i \in [n]} Y_i$. Then for any $t$ such that $nt>1$, there exists some constant $\constant$ such that:
    \begin{align*}
        \Pr[S_n > nt] &\leq \exp\left(-\frac{nt^2}{c\constant_x^2}\right) + \exp\left(-\frac{\sqrt{nt}}{4\constant_x}\right) + n\exp\left(-\frac{\sqrt{nt}}{2\constant_x}\right)
    \end{align*}
    Additionally, if $t \leq \frac{\constant_x^{2/3}}{\sqrt[3]{n}}$ and for sufficiently large $n$ ($n \geq \constant_x^2\log^3(n))$, then we have:
     \begin{align*}
     \Pr\left[\frac{1}{n} \sum_{i \in [n]} X_i^2 - \E[X^2] > t\right] \leq \constant_3 \exp\left(-\frac{nt^2}{\constant_x^2}\right)
     \end{align*}
    \begin{proof}
     Given $\Pr[|X_i| \geq t] \leq \exp\left(-\frac{t}{\constant_x}\right)$ with $\constant_x \geq 1$, for $Y_i = X_i^2 - \E[X_i^2]$, we have
        \begin{align*}
            \Pr[Y_i \geq t]    &= \Pr\left[|X_i| \geq \sqrt{t+ \E[X]^2}\right] 
            \leq \exp\left(\frac{-\sqrt{t+\E[X^2]}}{\constant_x}\right)
            \leq \exp\left(\frac{-\sqrt{t}}{\constant_x}\right)
        \end{align*}
        Where the third inequality follows from the fact that $\E[X^2]$ is a positive quantity.

        Then, $Y_i$ satisfies the conditions in Lemma \ref{Lemma: subweibull tail bound} with $\alpha = 2$, and $\constant_\alpha = -\frac{1}{\constant_x}$. We then fix $\beta = \frac{1}{2}$ to obtain
        \begin{align*}
            \Pr[S_n >  nt] \leq &\exp\left(-\frac{nt^2}{2\constant_x^2 + \frac{\constant_1}{\constant_x^4}+ (nt)^{-\frac{1}{2}}\frac{\constant_2}{\constant_x^5}}\right) + \exp\left(-\frac{\sqrt{nt}}{4\constant_x}\right) + n\exp\left(-\frac{\sqrt{nt}}{2\constant_x}\right)
        \end{align*}

        Using the values in the functions $\constant_1(\beta,\alpha)$ and $\constant_2(\beta,\alpha)$, we obtain  $\constant_1(\beta,\alpha) = \constant_1/\constant_x^4 \leq \constant_1$ and $\constant_2(\beta,\alpha) = \constant_2/\constant_x^5 \leq \constant_2$ where $\constant_1$ and $\constant_2$ are constants. Additionally using the fact that $nt > 1$, we obtain:
        
        \begin{align*}
            \Pr[S_n >  nt] \leq &\exp\left(-\frac{nt^2}{2\constant_x^2 + \constant_3}\right) + \exp\left(-\frac{\sqrt{nt}}{4\constant_x}\right) + n\exp\left(-\frac{\sqrt{nt}}{2\constant_x}\right)
        \end{align*}
         $\constant_x > 1$ ensures $\exp(-\frac{nt^2}{2\constant_x^2+c}) \leq \exp(-\frac{nt^2}{(c+2)\constant_x^2})$, hence we can rewrite the first inequality in the lemma.

         For the second part, note that the left side of the inequality is a direct consequence of putting $S_n = \sum_{i\in[n]} Y_i = \sum_{i\in[n]} X_i^2 - \E[X^2]$. For the right-hand side, under the assumption on the value of $n$, we have:
         \begin{align*}
             n &\geq \frac{1}{\constant}\constant_x^2\log^3(n)\\
             \implies n^{2/3} &\geq \frac{1}{\constant^{2/3}}\constant_x^{4/3}\log^2(n)\\
             \implies n^{2/3}t &\geq \constant'\constant_x^{2}\log^2(n)n^{-1/3} \\
             \implies nt &\geq \constant'\constant_x^{2}\log^2(n)\\
             \implies \frac{1}{\constant'}\sqrt{nt/\constant_x^2} &\geq \log(n)\\
             \implies \exp\left(-\frac{1}{\constant'}\sqrt{nt/\constant_x^2}\right)&\leq \exp\left(-\log(n)\right)
         \end{align*}
         This ensures that the last two terms on the right can be bounded as:
         \begin{align*}
             \exp\left(-\frac{\sqrt{nt}}{4\constant_x}\right) + n\exp\left(-\frac{\sqrt{nt}}{2\constant_x}\right) \leq \exp\left(-\frac{\sqrt{nt}}{\constant_1\constant_x}\right)
         \end{align*}
         for some positive $\constant_1$. Now, under the assumption on $t$, the first term on the right-hand side is larger than this bound:
         \begin{align*}
             t &\leq \constant\constant_x^{2/3}n^{-1/3}\\
             \implies t^{3/2} &\leq \constant\constant_x n^{-1/2}\\
             \implies \frac{nt^2}{\constant'\constant_x^2} &\leq \frac{\sqrt{nt}}{4\constant_x}\\ 
             \implies \exp\left(-\frac{nt^2}{\constant'\constant_x^2}\right) &\geq \exp\left(-\frac{\sqrt{nt}}{4\constant_x}\right)\,.
         \end{align*}
         By fixing the constants appropriately, the second term can be ensured to be larger than the third term on the right-hand side. This gives us the second inequality stated in the lemma.
    \end{proof}
\end{lemma}

To bound the left tail, we use the following well-known one-sided Bernstein style inequality for lower tails of non-negative random variables \cite{Wainwright_2019_HDS}:

\begin{lemma}[\textbf{One-Sided Bernstein Inequality}]\label{Lemma: One-Sided Bernstein}
    Given i.i.d. non-negative random variables $X_i$, we have:
    \begin{align*}
        \Pr\left[\sum_{i \in [n]} X_i - \E[X_i] \geq -nt\right] \leq \exp\left(-\frac{nt^2}{\E[X_i^2]}\right)
    \end{align*}
\end{lemma}

\begin{lemma}\label{lemma: Subexponential Squared Left Tail}
    Given an i.i.d. sequence  $X_i $ such that $\Pr[|X_i| \geq t] \leq \exp\left(-\frac{t}{\constant_x}\right)$ with $\constant_x \geq 1$, define a corresponding i.i.d sequence $Y_i = \E[X_i^2] - X_i^2$. Define $S_n = \sum_{i \in [n]} Y_i$. Then we have:
    \begin{align*}
        \Pr[S_n \geq nt] \leq &\exp\left(-\frac{nt^2}{\constant'_x}\right)
    \end{align*}

    \begin{proof}
        By the equivalent definitions of sub-exponential distributions, we have $\E[X_i^4] \leq \constant'_x$. Then, the Lemma \ref{Lemma: One-Sided Bernstein} gives us:
        \begin{align*}
            \Pr[S_n \geq nt] 
            =\Pr[X_i^2 - \E[X_i^2] \leq -nt]
            \leq \exp\left(-\frac{nt^2}{\constant'_x}\right)
        \end{align*}
        Here, the last inequality follows from Lemma \ref{Lemma: One-Sided Bernstein}.
    \end{proof}
\end{lemma}

Combining Lemma~\ref{lemma: Subexponential Squared Right Tail} and \ref{lemma: Subexponential Squared Left Tail}, we obtain the following both-sided bound for the square of Laplace random variables:

\begin{lemma}[\textbf{Tail Bounds for Squared Sub-Exponential Random Variables}]\label{Lemma: Laplace Squared Tail}
    Given an i.i.d. sequence $X_i $ such that $\Pr[|X_i| \geq t] \leq \exp\left(-\frac{t}{\constant_x}\right)$ with $\constant_x \geq 1$, define a corresponding i.i.d sequence $Y_i = X_i^2 - \E[X^2]$. Define $S_n = \sum_{i \in [n]} Y_i$. Then for any $t$ such that $t \leq \frac{\constant_x^{2/3}}{\sqrt[3]{n}}$ and for sufficiently large $n$ ($n \geq \constant_x^2\log^3(n))$, there exists some constant $\constant$ such that:
    \begin{align*}
     \Pr\left[\frac{1}{n} \sum_{i \in [n]} |X_i^2 - \E[X^2]| > t\right] \leq \exp\left(-\frac{\constant nt^2}{\constant_x^2}\right)
     \end{align*}
\end{lemma}

\begin{lemma}[\textbf{Deviation Bound for Laplace Random Matrix}]\label{lemma: laplace random matrix deviation bound}
    Let $\bX \in \R^{n\times d_1}$ and $\bY \in \R^{n\times d_2}$ be random matrices whose entries are random variables $X_{ij}$ and $Y_{ij}$ such that $\Pr[|X_{ij}| \geq t] \leq \exp\left(-\frac{t}{\constant_x}\right)$ and $\Pr[|Y_{ij}| \geq t] \leq \exp\left(-\frac{t}{\constant_y}\right)$ with $\constant_x ,\constant_y \geq 1$, , and let $\constant_{\max} = \max(\constant_x,\constant_y)$. then for any $t$ such that $t \leq \frac{\constant_{\max}^{2/3}}{\sqrt[3]{n}}$ and sufficiently large $n$ ($n \geq \constant_x^2\log^3(n))$:
    \begin{align*}
        \Pr\left[\norm{\frac{\bY^T\bX}{n}-\mathrm{cov}(y_i,x_i)}_{\max}> t\right] \leq d_1d_2\exp(-\frac{\constant_0\sqrt{nt}}{\constant_{\max}})
    \end{align*}
    Where $x_i$ and $y_i$ are the $i$-th rows of $\bX$ and $\bY$, respectively. Additionally, if $d = \max(d_1,d_2)$, 
    \begin{align}
        \Pr\left[\norm{\frac{\bY^T\bX}{n}-\mathrm{cov}(y_i,x_i)}_{\max} \geq \frac{\constant_1\constant_{\max}^2\log^2d}{n}\right] \leq \frac{1}{d^{\constant_2}}\label{eq: laplace covariance deviation bound}
    \end{align}
    Where, $\constant_0, \constant_1$ and $\constant_2$ are universal constants.
    \begin{proof}
        Observe that we can rewrite the $\norm{\cdot}_{\max}$ norm in terms of unit vectors $\be_i$ where only the $i$-th coordinate is $1$ and the rest are $0$.
        \begin{align*}
            \norm{\frac{\bY^T\bX}{n}-\mathrm{cov}(y_i,x_i)}_{\max} 
            = &\max_{\substack{(i,j) \in [d_1]\times[d_2]}}\be_i\left\{\frac{\bY^T\bX}{n} - \mathrm{cov}(y_i,x_i)\right\}\be_j \\ 
            = &\max_{\substack{(i,j) \in [d_1]\times[d_2]}} \frac{1}{2}[\Phi(\bX \be_j+\bY \be_i) - \Phi(\bX e_j)-\Phi(\bY \be_i)]
        \end{align*}
        Where $\Phi(\bv) = \frac{\norm{\bv}_2^2}{n} - \E\left[\frac{\norm{\bv}_2^2}{n}\right]$. Also, note that $\Pr[\bX \be_j+ \bY \be_i \geq t] \leq \exp(-\frac{t}{2\constant_{\max}})$ by union bound on either of $\bX \be_j$ or $\bY \be_i$ being $\geq \frac{t}{2}$. Hence, we have for $t \leq \frac{\constant_{\max}^{2/3}}{\sqrt[3]{n}}$:
        \begin{align*}
            \Pr\left[\be_i\left\{\frac{\bY^T\bX}{n} - \mathrm{cov}(y_i,x_i)\right\}\be_j \geq t\right] \leq \exp\left(\frac{-\constant nt^2}{\constant_{\max}^2}\right)
        \end{align*}
        And by a union-bound argument over all possible values of $i$ and $j$, we have:
        \begin{align*}
        \Pr\left(\norm{\frac{\bY^T\bX}{n}-\mathrm{cov}(y_i,x_i)}_{\max} \geq t\right)
        =&\Pr\left[\max_{\substack{(i,j) \in [d_1]\times[d_2]}} \be_i\left\{\frac{\bY^T\bX}{n} - \mathrm{cov}(y_i,x_i)\right\}\be_j \geq t\right]\\
        \leq& d_1d_2\exp\left(\frac{-\constant nt^2}{\constant_{\max}^2}\right)
        \end{align*}
        Fixing $t = \frac{\constant_1\constant_{\max}\sqrt{\log{d}}}{\sqrt{n}}$ satisfies $t \leq \frac{\constant_{\max}^{2/3}}{\sqrt[3]{n}}$ when $n \geq \constant\constant_{\max}^2\log^3(d)$. Hence, we obtain:
        \[
        \Pr\left[\norm{\frac{\bY^T\bX}{n}-\mathrm{cov}(y_i,x_i)}_{\max} \geq \constant_1\constant_{\max}\sqrt{\frac{\log{d}}{n}}\right] \leq \frac{1}{d^{\constant_2}}
        \]
    \end{proof}
\end{lemma}


\end{document}